\documentclass[review]{elsarticle}
\usepackage[hidelinks]{hyperref}
\usepackage{float}
\usepackage{verbatim} %comments
\usepackage{apalike}
\usepackage{color}
\usepackage[table,xcdraw]{xcolor}           % inserción de gráficos

\usepackage{pdflscape}
\usepackage{lscape}
\usepackage{tabularx}
\usepackage{amssymb}
\usepackage{amsmath}
\usepackage{caption}

\usepackage[normalem]{ulem}
\useunder{\uline}{\ul}{}

\usepackage{subfig}
\usepackage{varioref}
\usepackage[hidelinks]{hyperref}
\usepackage{cleveref}

\usepackage[ruled,lined,linesnumbered]{algorithm2e}

\SetCommentSty{mycommfont}

\newcommand{\x}{$\phantom{x}$}

\usepackage{graphicx}
\usepackage{xcolor,colortbl}
\usepackage{multirow}
\usepackage{algorithmic}

\usepackage{textcomp}
\usepackage[normalem]{ulem}

\definecolor{dgreen}{rgb}{0.0, 0.5, 0.0}

\usepackage{cleveref}
\crefformat{footnote}{#2\footnotemark[#1]#3}

\makeatletter
\def\ps@pprintTitle{%
 \let\@oddhead\@empty
 \let\@evenhead\@empty
 \def\@oddfoot{\centerline{\thepage}}%
 \let\@evenfoot\@oddfoot}
\makeatother

\begin{document}

\begin{frontmatter}

\title{MOELIGA: a multi-objective evolutionary approach for feature selection with local improvement}

\author[label1,cor1]{Leandro Vignolo}
\ead{ldvignolo@sinc.unl.edu.ar}

\author[label1]{Matias Gerard}
\ead{mgerard@sinc.unl.edu.ar}

\cortext[cor1]{Corresponding author.}
\address[label1]{Research Institute for Signals, Systems and Computational Intelligence (s{\i}nc(\textit{i})), FICH--UNL/CONICET, Ciudad Universitaria UNL, Santa Fe, Argentina.}

\begin{abstract}
Selecting the most relevant or informative features is a key issue in actual machine learning problems.
Since an exhaustive search is not feasible even for a moderate number of features, an intelligent search strategy must be employed for 
finding an optimal subset, which implies considering how features interact with each other in promoting class separability.
Balancing feature subset size and classification accuracy constitutes a multi-objective optimization challenge.
Here we propose MOELIGA, a multi-objective genetic algorithm incorporating an evolutionary local improvement strategy that evolves subordinate populations to refine feature subsets. MOELIGA employs a crowding-based fitness sharing mechanism and a sigmoid transformation to enhance diversity and guide compactness, alongside a geometry-based objective promoting classifier independence. 

Experimental evaluation on 14 diverse datasets demonstrates MOELIGA's ability to identify smaller feature subsets with superior or comparable classification performance relative to 11 state-of-the-art methods. These findings suggest MOELIGA effectively addresses the accuracy-dimensionality trade-off, offering a robust and adaptable approach for multi-objective feature selection in complex, high-dimensional scenarios.

\end{abstract}


\end{frontmatter}

%===========================================
\section{Introduction}
\label{introduction}
Current challenges in machine learning increasingly involve dealing with high-dimensional data. Technological advances enable collecting vast amounts of data, but in many applications, the number of samples cannot be increased proportionally \cite{cai2018}. This occurs particularly in biological applications, where only small representative datasets are available. In such scenarios, compact data representations are crucial. Predictive models built on low-dimensional representations exhibit less variability than those trained on original data, enhancing generalization and enabling effective predictions for new cases.

Two approaches address high-dimensional problems: dimensionality reduction and feature selection. Both facilitate modeling and interpretation of underlying phenomena. Dimensionality reduction transforms original attributes into a new space, eliminating feature interpretability \cite{mendes2020}. Moreover, the transformation criterion may not relate to discrimination, failing to improve classification performance. Feature selection (FS) identifies a subset of original features for the learning algorithm, preserving interpretability while potentially improving classification \cite{remeseiro2019}.

FS is challenging because relevant features are typically unknown. This makes FS essential for achieving classification with reduced complexity and acceptable performance \cite{song2024,cai2018}. FS selects attributes that efficiently describe data while reducing noise and irrelevant features without sacrificing prediction performance. Three common objectives guide this selection: maximizing accuracy with a specified subset size, finding the smallest subset satisfying an accuracy requirement, or optimizing the trade-off between dimensionality and accuracy. FS improves machine learning models in performance, computation time, complexity, interpretability, and generalization, while alleviating the curse of dimensionality.

FS techniques are categorized by evaluation criterion and search methodology \cite{guyon06}. Wrapper methods evaluate feature subsets using the predictive model, ensuring correlation with the metric of interest. However, training the model for each candidate subset makes evaluating all combinations infeasible. Metaheuristic search techniques address this computational challenge through stochastic optimization, finding good solutions with moderate cost. Examples include genetic algorithms (GAs), particle swarm optimization (PSO), and ant colony optimization (ACO) \cite{vieira12}. GAs are particularly suitable for FS due to binary encoding, while ACO and PSO use real-number representations.

% GAs are evolutionary algorithms that imitate natural evolution 
GAs are a type of evolutionary algorithms, a family of biologically inspired techniques that involve several mechanisms to imitate natural evolution \cite{katoch2021}. 
They have been successfully applied to feature subset optimization \cite{mendes2020,labani2020,vignolo13,hsu11,pedrycz12}, speech representations \cite{bakhshi2020, vignolo10}, speech recognition \cite{vignolo16}, face recognition \cite{ghouzali2020, vignolo13}, and microstructural image classification \cite{khan2021}. However, canonical GAs have limitations often neglected in FS applications \cite{eligaCLEI}. Solutions from classical heuristics, filter methods, and simple approaches are frequently comparable or superior to those from simple GAs.

Several approaches overcome classical GA limitations by incorporating do\-main-\-spe\-cif\-ic knowledge. Hybrid GAs use problem-specific encoding or specialized genetic operators \cite{vignolo16b,bui1996,ilseok04, song2024}. Recent proposals combine filter and wrapper strategies \cite{zhang2025} and integrate evolutionary computation with reinforcement learning and Markov chains \cite{rehman2024}. Other approaches employ local improvement mechanisms \cite{sivaram2019}, gradient descent refinement \cite{navarro12}, adaptive local search \cite{thangiah2019, Sinthamrongruk2017}, and embedded regularization \cite{liu2018}. Some methods exploit feature correlation \cite{kabir09}, though removing correlation does not always improve classification, as redundant information enhances robustness with noisy data \cite{vignolo10} and improves results in complex tasks \cite{vignolo16}.

% Single-objective methods use one quality function, forcing the population toward a particular feature set. 
Single-objective search methods force the population toward a particular zone of the
Pareto front due to the use of a single quality measure. Inappropriate objective functions may yield poor results, and single-objective methods often struggle to handle applications that require the optimization of competing objectives. Multi-objective optimization methods, like the multi-objective genetic algorithm (MOGA) \cite{konak2006}, naturally address this issue. Recent multi-objective evolutionary algorithms for FS include approaches with dominance-based initialization \cite{chen2025} and methods for time series forecasting that partition datasets and associate each partition with an objective function \cite{espinosa2025}.

% We propose MOELIGA, a novel FS method based on an improved multi-objective genetic algorithm (MOGA) \cite{konak2006}. 
% We propose MOELIGA, a novel Multi-Objective Lo\-cal\-ly-\-Imp\-roved  Genetic Algorithm for FS. The main contributions are: i) a local improvement strategy evolving subordinate populations in the active feature subspace of the best chromosomes; ii) a novel crowding-based fitness sharing technique promoting population diversity; iii) a second objective function with a sigmoid function making feature number changes relative to selected features; and iv) a third objective function based on nearest neighbour distances in the selected feature space preventing overfitting. Despite requiring more evaluations per generation than simple MOGAs, MOELIGA produces superior solutions in classification accuracy and feature dimensionality within fewer generations.

 In this work, we introduce MOELIGA, a novel Multi-Objective Locally-Improved Genetic Algorithm for feature selection. The main contributions of MOELIGA are: (i) a local improvement strategy that evolves subordinate populations within the active feature subspace of the best-performing chromosomes; (ii) a novel crowding-based fitness sharing technique to promote population diversity; (iii) a second objective function that uses a sigmoid transformation to make changes in the number of selected features more significant for smaller subsets; and (iv) a third objective function based on nearest neighbor distances in the selected feature space to help prevent overfitting. Although MOELIGA requires more evaluations per generation than standard MOGAs, it achieves superior classification accuracy and feature dimensionality trade-offs in fewer generations.

% The remainder is organized as follows. Section \ref{materialsandmethods} presents the materials and methods, including an overview of the multi-objective genetic algorithm. Section \ref{proposal} describes MOELIGA in detail, covering the evolutionary local improvement strategy, initialization procedure, objective functions, and fitness sharing mechanism. Section \ref{sec:results_and_discussion} reports and discusses results, including experimental setup, parameter analysis, comparisons with state-of-the-art methods, and performance evaluations across datasets and classifiers. Section \ref{Conclusions_and_future_work} concludes and outlines future work.

The remainder of this paper is organized as follows. Section \ref{materialsandmethods} presents the materials and methods, including an overview of the multi-objective genetic algorithm and a detailed description of MOELIGA, covering the evolutionary local improvement strategy, initialization procedure, objective functions, and fitness sharing mechanism. Section \ref{sec:results_and_discussion} reports and discusses the results, including the experimental setup, parameter analysis, comparisons with state-of-the-art methods, and performance evaluations across various datasets and classifiers. Section \ref{Conclusions_and_future_work} concludes the paper and outlines directions for future work.

%===========================================
% \section{Materials and methods}
\section{Fundamentals and proposal}
\label{proposal}

%-------------------------------------------
\subsection{Genetic algorithms for multi-objective problems}
\label{materialsandmethods}

Most optimization problems, formulated for practical applications, require multiple targets or requirements to be met. These can be regarded as multi-objective optimization problems, where each target is treated as a sub-problem \cite{xue21}. When dealing with these kinds of problems, a number of objective functions are simultaneously optimized. Then, the solution consists not in a single point, but a set of points known as the Pareto optimal set, each representing a different trade-off for a decision maker to choose~\cite{deb14}.
A general formulation for multi-objective optimization with no constraint, defining
each objective as a minimization function, is
\begin{equation}
\min F(\mathbf{x}) = \left( o_1(\mathbf{x}), o_2(\mathbf{x}), \dots , o_l(\mathbf{x}) \right), \phantom{XX} with \phantom{.} \mathbf{x} \in \Omega
\subseteq \mathbb{R}^N
\end{equation}
where $\mathbf{x} = (x_1, x_2, \dots, x_N)$ is a point in the $N$-dimensional search space $\Omega$ and $o_j(\mathbf{x})$ are $l$ objective functions defined in $F(\mathbf{x})$.
In real-life problems, coexisting objectives conflict with each other; therefore, the optimization
with respect to a single target often results in inadmissible results with respect to the other objectives.
This means that an ideal solution that simultaneously satisfies each objective entirely is infeasible for a challenging problem. In a 
minimization problem, a feasible solution $\mathbf{x}$ is said to dominate another feasible solution $\mathbf{y}$ if and only if 
$o_j(\mathbf{x})\leq o_j(\mathbf{y})$ $\forall j$ and $o_j(\mathbf{x})<o_j(\mathbf{y})$ for at least one $j$. 
The set of all feasible non-dominated solutions is referred to as the Pareto optimal set, and the corresponding objective function values for such solutions conform the Pareto front.
Then, the expected result of a multi-objective problem is a set of non-dominated solutions, each of which satisfies the objectives at 
an acceptable level.  
The goal in multi-objective optimization is to find solutions in the Pareto optimal set. 
However, identifying the entire Pareto optimal set is practically impossible in general,
and proving the optimality of solutions is computationally infeasible in combinatorial optimization. Therefore, a practical approach is to search for a set of solutions that represent the Pareto optimal set as well as possible, or \emph{the best-known Pareto set}.

Inspired by the natural process of evolution, the GA emerged as  meta-heuristic optimization methods,
capable of finding global optima in complex search spaces~\cite{katoch2021}.
The search procedure of GAs consists in evolving a population  of individuals, carrying the information 
of feasible solutions encoded in a binary chromosomes. 
Individuals compete with each other in order to produce offspring  based on their fitness,
imitating the selective pressure of a natural environment. The fitness is assessed by evaluating the objective functions on the solution decoded from their chromosomes.
In a classical GA, the chromosomes are manipulated by mechanisms of selection, mutation, crossover and replacement \cite{xue21}. The selection operator plays as the survival-of-the-fittest mechanism in nature, so that the chance of an individual to survive is proportional to its fitness. Progenitors are selected from the population on every generation, in order to produce offspring by means of the variation operators (mutation and crossover). The purpose of the variation operators is to combine information from different individuals (crossover) and also to maintain population diversity, by randomly modifying chromosomes (mutation).
A replacement strategy is then applied in order to replace, partially or completely, the current population by the offspring. These steps are repeated until a desired termination criterion is reached; for example, a predefined number of generations or a desired fitness value \cite{khan2021,katoch2021,xue21}.

%-------------------------------------------
\subsection{Evolutionary local improvement}
\label{proposal:evolutionary_local_improvement}

We introduce a variation of the MOGA where selected solutions undergo local improvement. This strategy extends previous work on 
genetic algorithms with aggregative fitness functions \cite{eligaCLEI}, which outperformed other methods. The local improvement 
strategy evolves subordinate populations based on chromosomes selected from the main population. Each subordinate population is 
generated from a selected chromosome used as a template. New individuals are created as random combinations of the corresponding 
features and evolved using another MOGA. In this embedded optimization, fitness is evaluated as in the main population, with the first 
objective function assessing classification using only the features selected in the chromosome.

The pseudocode of the proposed \emph{multi-objective genetic algorithm with evolutionary local improvement}
(MOELIGA) is presented in Algorithm \ref{pseudo:MOGA}. 
Subordinate populations are evolved in a similar way to the main population, and the number of generations in which these are evolved is added to the number of generations for the main population (lines 9 to 13 in Algorithm \ref{pseudo:MOGA}).
Once the stop criterion is met in the subordinate evolution loop, a replacement strategy is applied to update the main population (line 14 in Algorithm \ref{pseudo:MOGA}). 
We propose and compare three different replacement strategies: parent replacement (PR), complete replacement (CR) and selection replacement (SR).

In the PR %parent replacement 
strategy, the fittest individual for each subordinate population is compared with the parent template, which is replaced in the main population if its performance is improved (i.e. either  the classification accuracy or the number of features is improved without degradation of the other).
In the CR %complete replacement 
strategy, all individuals from the main population and from all subordinate populations are gathered and sorted according to their fitness. From these, the best $N_{p}$ individuals are kept to form the new main population, with $N_{p}$ being the size of the original main population.
As in the previous case, the SR %selection replacement
strategy brings together individuals from the main population and subordinate populations. But here, the $N_{p}$ individuals that will be part of the new main population are picked by means of the selection operator.

\begin{figure}
    \centering
    \includegraphics[width=\textwidth]{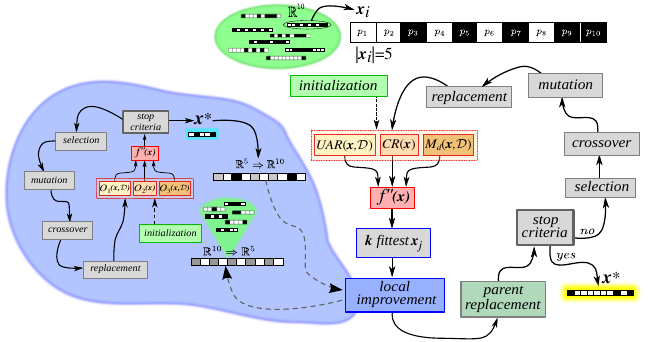}
    \caption{Approach with evolutionary local improvement for multi-objective feature selection.}
    \label{fig:moeliga}
\end{figure}

Figure \ref{fig:moeliga} outlines the MOGA with the proposed local evolutionary improvement strategy. The main MOGA runs over the full feature space, and a subordinate MOGA performs the local optimization of the fittest individuals (blue area in the figure). In this example, the dataset contains 10 features ($p_{1}$ to $p_{10}$), and chromosomes represent different feature subsets. The local improvement strategy takes the \textit{k} fittest chromosomes (parent chromosomes) from the main population to create subordinate populations in order to improve the corresponding subsets of features (the figure shows 5 selected features as an example). Thus, the chromosomes in a subordinate population represent different combinations of the features in this subset. The subordinate MOGA is run on this new feature space, evaluating the population with the same objective functions and dataset as in the main MOGA.
After the subordinate population is evolved, the best chromosome 
(2 features or black boxes in the figure) is compared with the corresponding parent chromosome from the main MOGA, which is replaced in the main population if the fitness is improved (PR strategy).

%  TASA DE INICIALIZACIÓN ESCALONADA
\subsection{Staggered Initialization}
\label{sec:init}
In MOELIGA the chromosomes encode subsets of features by means of binary codification, where the $1$ value bits indicate active or 
selected features. Small feature subsets are preferred, and we look for diversity in the population not only regarding the selected 
attributes but also the proportion of active genes.
For this reason, we introduced a staggered initialization scheme in which the population is divided into three
parts, and the proportions of genes that are randomly set to $1$ are different for each part.
The first part of the population, containing the $55\%$ of the chromosomes, is initialized with $3\%$ of active genes; the second part, containing $30\%$ of the chromosomes, is initialized with $15\%$ of active genes, and the remaining $15\%$ of the chromosomes are initialized with $35\%$ of active genes.
This way, it is guaranteed that the initial population contains chromosomes with different numbers of selected attributes, and most individuals contain a relatively small number of active genes.

\begin{algorithm}[t!]
\DontPrintSemicolon
% \dontprintsemicolon
\LinesNumbered
% \linesnumbered
{\small
    
    Initialize the GA population\; 
    \textbf{Evaluate population} (Algorithm \ref{algo:evaluate}) \; 
    \Repeat{stopping criteria is met}{
        Parent selection\;        
        Mate selected parents (crossover)\;
        Mutate  offspring \;
        Replace population\;
        \textbf{Evaluate population} (Algorithm \ref{algo:evaluate})\;
        \tcc{Local improvement of best chromosomes}
        \ForEach{\emph{of the $N_b$ best individuals}}{
               
             Create template chromosome based on selected genes\;
             Create subordinate population based on template\;
             Evolve subordinate population by multi-objective GA (1 to 8)\;
        }
        Replace improved chromosomes in the main population\;
     }
}
\caption{GA with evolutionary local improvement}
\label{pseudo:MOGA}
\end{algorithm}

\subsection{Objective functions}

\subsubsection{Objective I} 
\label{proposal:uar}

The goal of the feature selection process is to maximize classification accuracy while reducing the number of attributes. Many machine learning datasets show unbalanced class distributions, leading to over-estimation of classifier performance. To avoid this issue,  classification performance is measured using the \emph{Unweighted Average Recall} (UAR)\cite{Rosenberg12}, also known as \emph{Balanced Accuracy}, which is calculated as:
\begin{eqnarray}
	UAR(\mathbf{x},\mathcal{D}) & = & \frac{1}{C_\mathcal{D}} \displaystyle\sum_{i=1}^{C_\mathcal{D}} RC_{i}(\mathbf{x},\mathcal{D}), \label{eq:uar}\\
RC_{i}(\mathbf{x},\mathcal{D}) & = & \frac{TP_i(\mathbf{x},\mathcal{D})}{TP_i(\mathbf{x},\mathcal{D})+FN_i(\mathbf{x},\mathcal{D})}    	
\end{eqnarray}
\noindent where
$RC_{i}(\mathbf{x},\mathcal{D})$ is the recall obtained for $ith$-class using the features selected by chromosome $\mathbf{x}$, $C_\mathcal{D}$ is the number of classes in dataset $\mathcal{D}$, $TP_i(\mathbf{x},\mathcal{D})$ is the number of instances of class $i$ correctly classified, and $FN_i(\mathbf{x},\mathcal{D})$ is the number of examples of class $i$ incorrectly classified. This measure takes into account the number of instances for each class and weighs the recall accordingly. This makes the UAR metric insensitive to skewed datasets, meaning that it is not affected by the imbalance between the number of examples from each class.

For the evaluation of a particular chromosome, the UAR is computed based on a validation dataset, considering only the features selected by the chromosome. This means that the classifier is trained and evaluated on the dataset
using selected features only.
In order to avoid overfitting of the feature selection process, the evaluation of the classifier is repeated $N$ times, with a different random train/validation split each time. In this way, the classifier is tested with different instances each time, and the average UAR is assigned to Objective 1. 
The steps involved in the computation of the fitness value for each individual are summarized in Algorithm \ref{algo:evaluate} from line 1 to 9.

\subsubsection{Objective II} 
\label{proposal:nfeatures}
In order to account for the number of selected features as an objective function, we compute the cardinality ratio of a given 
chromosome $\mathbf{x}$ as
\begin{equation}
 \textrm{CR}(\mathbf{x}) = \frac{N_F-N_s(\mathbf{x})}{N_F}, 
\label{eq:featureratio}
\end{equation}

\noindent where $N_F$ is the size of the chromosome (i.e. the number of available features) and $N_s$ is the number of selected features.
Also, to control the relative impact of the number of features, we defined an alternative function that performs a non-linear mapping 
\begin{equation}
\textrm{CR}_\lambda(\mathbf{x}) = \frac{1}{1 + e^{-\lambda\left[  \textrm{CR}(\mathbf{x}) + \gamma\right]}}
\label{eq:CRlambda}
\end{equation}

\noindent where sigmoid parameters $\lambda$ and $\gamma$ are set 
to control the mapping, making the difference between solutions with a smaller number of features more significant than that of solutions with a larger number of features. For example, the non-linearity allows that two chromosomes with $10$ and $15$ features receive significantly different $\textrm{CR}_\lambda$ values, while two chromosomes with $1000$ and $1005$ features receive close $\textrm{CR}_\lambda$ values for the same feature selection problem.

%%%%%%%%%%%%%%%%%%%%%%%%%%%%%%%%%%%%%%%%%%%%%%%%%%%%%%%%%%%%%%%%%
\subsubsection{Objective III}
\label{subsec:obj3}
A third objective function is considered in this work to mitigate overfitting and to promote classifier independence in the feature selection process. This objective function relies on a distance-based score computed in the feature space, leveraging the concept of nearest neighbors to evaluate the quality of a feature subset. Specifically, for each instance in the dataset, two key references are identified: the \textit{nearestHit}, defined as the closest instance belonging to the same class, and the \textit{nearestMiss}, defined as the closest instance from a different class.

To compute this score for a given chromosome $\mathbf{x}$, a total of $N_{i}$ instances are randomly sampled from the dataset $\mathcal{D}$; each sampled instance is denoted by $\mathbf{d}^{(i)}$. The notation $\mathbf{d}_{\mathbf{x}}^{(i)}$ indicates that the instance $\mathbf{d}^{(i)}$ is represented using only the features selected by the solution $\mathbf{x}$. The distance-based score is then computed as
\begin{equation}
M_{d}(\mathbf{x},\mathcal{D})= \frac{1}{N_n}\sum_{i=1}^{N_n} \bigg( \frac{\Vert\mathbf{d}_{\mathbf{x}}^{(i)}-nearMiss\Vert_{\ell_1}}{\mid\mathbf{d}_{\mathbf{x}}^{(i)}\mid} - \frac{\Vert\mathbf{d}_{\mathbf{x}}^{(i)} -nearHit\Vert_{\ell_1}}{\mid\mathbf{d}_{\mathbf{x}}^{(i)}\mid} \bigg),
\label{mdist}
\end{equation}

\noindent where $\Vert \cdot \Vert_{\ell_1}$ denotes the $\ell_{1}$-norm of the argument, and $\mid \cdot \mid$ indicates the cardinality of the feature subset (i.e., the number of selected features). Both distance terms are normalized by the cardinality to ensure comparability across feature subsets of different sizes.

The interpretation of this metric is straightforward: higher values of  $M_{d}(\mathbf{x},\mathcal{D})$ indicate better feature subsets from the perspective of class separability. Specifically, the score increases when instances become closer to their \textit{nearestHit} (reducing the second term) and simultaneously farther from their \textit{nearestMiss} (increasing the first term). Conversely, the score decreases when instances are distant from same-class neighbors or close to different-class neighbors, which indicates poor discriminative power of the selected features. The complete procedure for computing the score for an arbitrary chromosome $\mathbf{x}$ is presented in Algorithm \ref{algo:distMetric}.

\begin{algorithm}[t!]
\DontPrintSemicolon
% \dontprintsemicolon
{\small
    \KwData{$\mathcal D$,$\phantom{.}N_i$,$\phantom{.}\mathbf{x}$}

     Initialize metric $M_{dist}=0$\;
     \For{$i=1$ to $N_n$}{
           Randomly sample $\mathbf{d}^{(i)} \sim U(\mathcal{D})$ with replacement\;
           Consider instance $\mathbf{d}^{(i)}$ using the features selected in $\mathbf{x}$, $\mathbf{d}_\mathbf{x}^{(i)}$\;
           Find the $nearestHit$ and the $nearestMiss$ of $\mathbf{d}_\mathbf{x}^{(i)}$ in $\mathcal{ D}_\mathbf{x}$\;
             $M_{d}= \displaystyle M_{d} +  \frac{\Vert\mathbf{d}_\mathbf{x}^{(i)} - 
nearMiss\Vert_{\ell_1}}{\mid\mathbf{d}_\mathbf{x}^{(i)}\mid}
             - \frac{\Vert\mathbf{d}_\mathbf{x}^{(i)} - nearHit\Vert_{\ell_1}}{\mid\mathbf{d}_\mathbf{x}^{(i)}\mid}\phantom{.....}$ (Eq. \ref{mdist})\;
       }
       Return $\displaystyle \frac{M_{d}}{N_n}$\;
}
\caption{Computation of distance metric used as 3rd objective.}
\label{algo:distMetric}
\end{algorithm}

Particularly, for binary feature vectors the distance metric $M_{d}(\mathbf{x},\mathcal{D})$ is bound\-ed in the range $[-1,1]$. The upper bound $M_{d}=1$ represents perfect class sep\-a\-ra\-bil\-i\-ty, where instances from different classes differ in all selected features while instances from the same class are identical. Conversely, $M_{d}=-1$ indicates that the selected features capture intra-class variability while failing to discriminate between classes. The value $M_{d}=0$ suggests that the selected features provide no discriminative power, as the average inter-class distance equals the average intra-class distance. This bounded range ensures direct comparability across feature subsets of different cardinalities, facilitating multi-objective optimization.

%================================================================

\subsection{Fitness sharing}

Let $\rm P_{g}$ denote the population at generation $\rm g$. To evaluate the fitness of the population, a rank $\rm r(\mathbf{x},g)$ is first assigned to each solution $\mathbf{x} \in \rm P_{g}$ according to
\begin{equation}
\rm r(\mathbf{x},g) = 1 + \rm \mathrm{nd}(\mathbf{x},g),
\label{eqrank}
\end{equation}

\noindent where $\rm nd(\mathbf{x},g)$ is the number of individuals in the generation $\rm g$ that dominate solution $\mathbf{x}$. This ranking assigns lower ranks to non-dominated solutions and higher ranks to dominated solutions, penalizing individuals in dominated regions of the objective space. Fitness is then computed based on rank as

\begin{equation}
 f(\mathbf{x}, \rm{g}) = N - \sum_{k=1}^{\rm r(\mathbf{x},g)-1} n_k - \frac{\rm n_{r(\mathbf{x},g)}-1}{2},
 \label{eqfit1}
\end{equation}

\noindent where $N$ is the population size, $\rm n_k$ is the number of individuals with rank $\rm k$, and $n_{r(\mathbf{x},g)}$ is the number of individuals sharing the same rank as $\mathbf{x}$. The first term ensures positive fitness values. The summation reduces fitness based on individuals in superior ranks, while the final term implements fitness sharing among individuals with equal rank, promoting diversity.

For instance, consider a population of $N=10$ individuals where the solution $\mathbf{x}$ has rank $r(\mathbf{x},g)=3$. Suppose there are $n_1=2$ individuals with rank 1 (non-dominated solutions), $n_2=3$ individuals with rank 2, and $n_3=4$ individuals with rank 3 (including $\mathbf{x}$ itself). The fitness is calculated as $f(\mathbf{x},g) = 10 - (2+3) - (4-1)/2 = 10 - 5 - 1.5 = 3.5$. The summation term $(2+3)=5$ accounts for all individuals in superior ranks (ranks 1 and 2), penalizing $\mathbf{x}$ for being dominated. The fitness sharing term $(4-1)/2=1.5$ reduces fitness further because $\mathbf{x}$ shares its rank with three other solutions, encouraging exploration of less crowded regions.

To further promote diversity and better exploration of the search space, we employ a fitness sharing technique that penalizes individuals based on their proximity to others. Good solutions in densely populated regions receive lower fitness than comparable solutions in less populated areas. This requires computing the Euclidean distance between every pair of solutions $\mathbf{x}$ and $\mathbf{y}$, which can be calculated in either the normalized objective space or the decision variable space. In the objective space, considering $N_o$ objective functions, the distance is

\begin{equation}
 d_o(\mathbf{x},\mathbf{y}) = \sqrt{ \sum_{j=1}^{N_o} \left( \frac{o_j(\mathbf{x})-o_j(\mathbf{y})}{o_j^{max}-o_j^{min}} \right)^2 },
 \label{distO}
\end{equation}

\noindent where $o_j^{max}$ and $o_j^{min}$ are the maximum and minimum observed values of the objective function $o_j(.)$, respectively. Alternatively, the distance in the decision variable space is

\begin{equation}
 d_x(\mathbf{x},\mathbf{y}) = \sqrt{ \frac{1}{N}\sum_{l=1}^N \left( x_l - y_l \right)^2 },
 \label{distX}
\end{equation}

\noindent where $N$ is the number of dimensions or, equivalently, the chromosome length.

To obtain the \emph{shared fitness} for each solution, we propose a crowding criterion that accounts for clusters of individuals concentrated in different search space regions. The \emph{shared fitness} $f'(\mathbf{x},\rm{g})$ of the solution $\mathbf{x}$ in generation $g$ is obtained by penalizing $f(\mathbf{x}, \rm{g})$ based on the clusters containing $\mathbf{x}$. For each generation, clusters are determined using a \emph{sharing function} $S$, defined as
\begin{equation}
\mathit{S}(\mathit{d}_\ast(\mathbf{x},\mathbf{y}),\sigma,\alpha) = 1 - \left[ \frac{\mathit{d}_{\ast}(\mathbf{x},\mathbf{y})}{\sigma}\right]^\alpha,
\label{sharingfun}
\end{equation}

\noindent with parameters $\sigma$ and $\alpha$ controlling the niche size and taper, respectively. 
These parameters allow for easy adjustment of the sharing function to emphasize penalization in crowded areas. Here $\mathit{d}_\ast$ is replaced with any of the distance functions defined in equations (\ref{distO}) and  (\ref{distX}).
We use this alternative in our work because preliminary experiments showed that the quality of the non-dominated solution sets is similar to that produced by the original sharing function presented in \cite{konak2006}, with the advantage of making it easier to adjust the penalization of populated areas of the search space.

The sharing function $S$ determines whether two individuals belong to the same cluster based on their distance $\mathit{d}_{\ast}$. An individual can belong to multiple clusters, with $S(\mathit{d}_\ast(\mathbf{x},\mathbf{y}),\sigma,\alpha)>0$ for all $\mathbf{x}$ and $\mathbf{y}$ in the same cluster. The \emph{shared fitness} for the solution $\mathbf{x}$ is then computed by penalizing its fitness based on proximity to the centroid of each cluster $j$ containing $\mathbf{x}$:

\begin{equation}
\mathit{f}'(\mathbf{x},\rm{g})=\sum_{\substack{\phantom{\sum}j\phantom{\sum}\\ \mathbf{x}\phantom{.}\in\phantom{.}\textrm{Cluster}_j}} \mathit{f}'(\mathbf{x},\rm{g})\left[1 - \mathit{S}(\mathit{d}_\ast(\mathbf{x},\mathit{Centroid_j}),\sigma,\alpha) \right]
\label{eqsharedfit} 
\end{equation}

\noindent with $\mathit{f}'(\mathbf{x},\rm{g})$ initialized as $\mathit{f}(\mathbf{x},\rm{g})$. Note the recursion on $\mathit{f}'(\mathbf{x},\rm{g})$, as it is penalized according to the distance from all cluster centroids containing $\mathbf{x}$.

Finally, the definitive fitness $f''(\mathbf{x}, \rm{g})$ for $\mathbf{x}$ in generation $\rm{g}$ is obtained by normalizing the fitness by the ratio of its shared fitness to the accumulated shared fitness of all solutions $\mathbf{y}$ with the same rank:

\begin{equation}
 f''(\mathbf{x}, \rm{g})= \mathit{f}(\mathbf{x},\rm{g}) \cdot  \frac{\mathit{f'}(\mathbf{x},\rm{g})}{\sum\limits_{\substack{\mathbf{y}\in \rm P_g\\ \rm r(\mathbf{y},g)=r(\mathbf{x},g)}} \mathit{f'}(\mathbf{y},\rm{g})}.
 \label{eqnormasfit}
\end{equation}

%%%%%%%%%%%%%%%%%%%%%%%%%%%%%%%%%%%%%%%%%%%%%%%%%%%%%%%%%%%%%%%%%%%%%%%%%%%%%%%%%%

\subsection{Fitness evaluation}

To compute the fitness of the population in MOELIGA, the
objective functions are first evaluated for each chromosome $\mathbf{x}$.
The first objective function in MOELIGA involves the evaluation of a
classifier to obtain the classification metric. For this purpose,
every instance in the dataset is represented using the attributes or features
selected by $\mathbf{x}$. Then, the classifier is trained and tested using the
re-parameterized dataset, and, to avoid overfitting of the feature selection 
process, this process is repeated $N$ times. At each iteration, a random stratified split
of the re-parameterized dataset is generated so that the classifier is tested with
different instances each time. The average of the test classification metric is then 
assigned to Objective 1. 
Then, the metric given in Eq. \ref{eq:CRlambda} is computed as Objective 2,
and the output of Algorithm \ref{algo:distMetric} is assigned as Objective 3.
This procedure is detailed step-by-step in the first \emph{for loop} (lines 1 to 12) 
of Algorithm \ref{algo:evaluate}. 

In order to determine how many chromosomes dominate a particular chromosome $\mathbf{x}$, 
the objective values for all the chromosomes in the current population are required. 
This is why the \emph{rank}, the \emph{fitness} and the \emph{shared fitness}
are computed in a separate \emph{for loop} (lines 13 to 18) in Algorithm \ref{algo:evaluate}.

Finally, once the \emph{shared fitness} value has been computed for the entire population,
the normalized fitness (Eq. \ref{eqnormasfit}) for every chromosome is obtained 
based on these values in the last \emph{for loop} of Algorithm \ref{algo:evaluate}.

\begin{algorithm}[t!]
\DontPrintSemicolon
% \dontprintsemicolon
\LinesNumbered
% \linesnumbered
{\small
     \tcc{Iterate over all chromosomes to evaluate the
     objective functions}
     \For{each chromosome $\mathbf{x}$ in the population}{
         Determine feature subset based on $\mathbf{x}$\;
         Re-parametrize data using the feature subset\;
         \For{each of $N_T$ tests}{
           Perform random stratified split of dataset\;
           Train the classifier on the training set\;
           Test the classifier on the validation set\;
         }
         Assign average test UAR to Objective 1 (Eq. \ref{eq:uar})\;
         Assign cardinality-ratio to Objective 2 (Eq. \ref{eq:CRlambda})\;
         Compute class separability metric (Algorithm \ref{algo:distMetric}) on the training and assign to Objective 3\;
     }
     \tcc{Iterate over all chromosomes to compute shared fitness,
     based on the objective values computed for the  population}
     \For{each chromosome $\mathbf{x}$ in the population}{    
     Assign $\mathbf{x}$ a Rank based on Pareto dominance (Eq. \ref{eqrank})\;  
     Compute the fitness for $\mathbf{x}$ based on it's rank (Eq. \ref{eqfit1})\; 
     Find all the clusters containing $\mathbf{x}$ based on Eq. \ref{sharingfun}\;
     Compute the Shared Fitness for $\mathbf{x}$ using Eq. \ref{eqsharedfit}\;       
     }
     \tcc{Iterate over all individuals in order to normalize the fitness
     based on the shared fitness of the entire population}
     \For{each individual in the population}{
     Normalize Fitness based on Shared Fitness (Eq. \ref{eqnormasfit})\;
     }
}
\caption{Evaluate population}
\label{algo:evaluate}
\end{algorithm}

\section{Results and discussion}
\label{sec:results_and_discussion}

Performance of methods in this work were evaluated using 14 datasets. 
Table \ref{tbl:datasets} summarizes the relevant meta-data for each of 14 datasets used, which correspond to two-class (3 
datasets) and multi-class (11 datasets) classification tasks, and involve a wide range of features (from 14 to 20000). Furthermore, 
these can be grouped in four sets: (i) synthetic dataset (\textit{Madelon}); (ii) general real-world classification tasks; (iii) 
handwritten digits classification; (iv) cancer-type classification.

%====================================================================================
 \begin{table}[H]        
 \centering
    \caption{Summary of the datasets used in the experiments.}    
    \label{tbl:datasets}
    \fontsize{9pt}{9pt}\selectfont
    \addtolength{\tabcolsep}{-0.3em}
    \begin{tabularx}{\hsize}{p{18mm} c c c p{6.7cm}}

\hline
\hline               
            & Samp. & Feat. & Class.  & \multicolumn{1}{c}{Description} \\
\hline 
    madelon              &   2400   &   500       &  2            & Classification of artificial data\footnotemark[1] 
\cite{bib_madelon}. \\    
    \hline
    dermatology &    358\footnotemark[2]   &    34       &  6   & Classification of Eryhemato-Squamous Disease type from 
biopsy and clinical  features \cite{bib_dermatology}. \\
    movement             &    360   &    90       & 15            & Classification of gestures in sign language (LIBRA) based on hand 
movements \cite{bib_movement}. \\
    arrhythmia           &    452   &   279       & 13            & Classification of arrhythmia events from ECG data 
\cite{bib_arrhythmia}. \\
    smartphone-activity	 &  10299   &   561       &  6            & Classification of smartphone activity from time and frequency 
domains \cite{bib_smartphone_activity}. \\
    isolet               &   7797   &   617       & 26            & Classification of spoken letter-name from speech features 
\cite{bib_isolet}. \\
    
    \hline
    optdigits	         &   5620   &    64       & 10            & Classification of handwritten digits  from pixel data 
\cite{bib_optdigits}. \\
    mfeat                &   2000   &   649       & 10            & Classification of handwritten digits based on features extracted 
from images \cite{bib_optdigits}. \\
    gisette              &   7000   &  4999       &  2            & Classification of handwritten digits $4$ and $9$ from image pixel 
data \cite{bib_gisette}. \\
    
    \hline
    leukemia             &     72   &  7129       &  2            & Classification of acute myeloid and lymphoblastic leukemia from 
gene expression\footnotemark[3] \cite{bib_leukemia}. \\
    all-leukemia         &    327   & 12558       &  7            & Classification of pediatric acute lymphoblastic leukemia genetic 
subtypes from microarrays \cite{bib_all_leukemia}. \\
    yeoh                 &    248   & 12625       &  6            & Classification of six pediatric lymphoblastic leukemia subtypes 
from gene expression \cite{bib_yeoh}. \\
    GCM                  &    327   & 16063       & 14            & Multiclass diagnosis of 14 tumor types from gene expression 
microarrays \cite{bib_gcm}. \\
    tcga-pancan          &    801   & 20531       &  5            & Classification of five cancer types from TCGA RNA-Seq gene 
expression \cite{bib_tcga_pancan}. \\
\hline 
    \end{tabularx}
    \end{table}

% :::::::::::::::::::::::::::::::::::::::::::

\footnotetext[1]{\texttt{https://archive.ics.uci.edu}}
\footnotetext[2]{366 total instances, 358 instances without missing values.}
\footnotetext[3]{table footnote 3}
%====================================================================================

\subsection{Experimental setup}

The proposed algorithm was implemented in C++11 (GCC 11.3.1), utilizing the MPICH2 message-passing library (version 1.1.1p1) for multi-processor and multi-core parallelization, and the mlpack library (version 3.2.2-3) for classification algorithms. The complete source code and instructions for MOELIGA are available on Github\footnote{\url{https://github.com/ldvignolo/moeliga}}.  For the implementation of other state-of-the-art approaches, Python 3.8.10 was used along with utilities from scikit-learn (version 0.24.1), and feature selection algorithms from skrebate (0.61), mlxtend (0.18.0), and boruta (0.3). Parallelization was achieved using the multiprocessing (2.6.0.2) and joblib (0.16.0) libraries. All experiments were conducted on a workstation equipped with an AMD Ryzen 7 1800X 3.60GHz CPU and 64GB of RAM.

\subsection{Basic parameters}

In the experiments, the effect of the parameters introduced with our approach
was evaluated, while other common GA parameters were fixed based on 
preliminary experiments and previous works \cite{eligaCLEI}.

The size of the main population was set to $90$ individuals, which were randomly initialized 
as described in Section \ref{sec:init}. In every generation, the population was replaced according to an elitist strategy, in which the best 10 individuals were maintained, together with a generational gap of the other 10 individuals, which were picked by the selection mechanism.
For the selection of individuals to create the offspring and the generational gap, we used the well-known roulette wheel method \cite{tang96}. The crossover and mutation rates were set to $0.9$ and $0.15$, respectively. As stop criteria, we considered a maximum of $300$ generations for all the experiments. 

The subordinate populations were of size $50$, and they were evolved for $70$ generations. The number of subordinate populations was $3$, which were created and evolved once every five generations of the main population. 
It must be noted that the evolution of the main population is stopped until the subordinate population has completed $70$ generations. In order to maintain computational complexity as low as possible, subordinate populations were evolved once every five generations of the main population.

To evaluate the accuracy of candidate solutions, we used a classifier based on decision trees with a maximum depth of $100$ \cite{kotsiantis2013}.
This classifier was chosen for the experiments because it achieves satisfactory performance for all the datasets considered with low computational cost. For this classification algorithm
we used the implementation provided in \emph{MLPack} \cite{curtin2013}, an efficient machine learning library for C++.
 
 For the evaluation of individuals during the optimization,  the classifier was trained using a \emph{training} data set and tested on a \emph{validation} set for the assessment of the accuracy of candidate solutions. 
 In order to avoid overfitting in the feature selection process, the training set was randomly split on every evaluation in order to set aside $30$\% of the instances for the validation set.  Once the search for the best feature subset was finished, the generalization performance was evaluated with the same classifier and a separate \emph{test} set.

%===========================================

\subsection{Sensitivity Analysis of MOELIGA Hyperparameters}
\label{sec:hyperparameter_effect_study}

This section analyzes the effect of the most relevant hyperparameters on the performance of MOELIGA. We use \textit{experiment} term to denote a specific hyperparameter configuration. Each experiment comprises multiple \textit{replications}, which are independent executions of the algorithm with identical hyperparameters but different random initializations. Statistics derived from these replications characterize the performance of each configuration.

To evaluate performance, we must first establish a method for assessing the solutions returned by MOELIGA. The algorithm outputs a set of non-dominated solutions approximating the true Pareto front, each representing a different trade--off between objectives. To enable comparison with state-of-the-art FS methods, we define a selection criterion for identifying a single representative solution from this set, balancing all objectives. This criterion is given by

\begin{equation}
R_{1}(\mathbf{x},\mathcal{D})=1 - \sqrt{{( 1 - UAR(\mathbf{x},\mathcal{D}) ) }^2 + {( 1 - \textrm{CR}_\lambda(\mathbf{x}) ) }^2},
\label{eq:R1_nolineal}
\end{equation}

\noindent where $UAR(\cdot)$ and $\textrm{CR}_\lambda(\cdot)$ correspond to the objectives I and II described by Eqs. (\ref{eq:uar}) and (\ref{eq:CRlambda}), respectively. This measure assigns each solution a grade balancing classification performance and feature subset size based on the Euclidean distance to the ideal point $(1,1)$. Solutions closer to this ideal point represent better trade-offs between objectives. These selected individuals characterize each experiment and enable comparisons between experimental configurations and with other algorithms.

To analyze the effect of MOELIGA hyperparameters on performance, a two-phase study was conducted. The first phase consisted of a \emph{screening} stage evaluating $7$ hyperparameters across all possible value combinations, resulting in $318$ experiments with $5$ replications each. Due to computational constraints, only three datasets were considered: GCM, Madelon, and Leukemia, representing problems with low, medium, and high feature dimensionality, respectively. The second phase performed a \emph{refinement} focusing on hyperparameters whose impact remained inconclusive from the screening phase. This second phase evaluated $10$ datasets, testing the selected hyperparameters over expanded value ranges with 5 replications per configuration. Screening and refinement phases are explained in detail in the following subsections.

To assess solution quality in these experiments, we employed a modified version of Eq. \ref{eq:R1_nolineal} that considers feature count without the nonlinear transformation. The nonlinearity in Eq. \ref{eq:R1_nolineal} was originally introduced to scale the importance of feature reduction relative to the ratio between selected and total features, which is essential for guiding the search process. However, this transformation is unnecessary when comparing final solution sets, as the linear relationship provides a more direct performance assessment. This modified function is defined as:

\begin{equation}
    \hat{R_{1}}(\mathbf{x},\mathcal{D}) = 1 - \sqrt{(1 - \textrm{UAR}(\mathbf{x},\mathcal{D}))^{2} + \left( 1 - \textrm{CR}(\mathbf{x}) 
\right)^{2}}
    \label{eq:R1}    
\end{equation} 

\noindent where $\textrm{UAR}(\cdot)$ represents the classification performance evaluated using Eq. \ref{eq:uar} on the test dataset considering only selected features; and $\textrm{CR}(\cdot)$ (Eq. \ref{eq:featureratio}), accounts for the features selected by chromosome $\mathbf{x}$ without using the nonlinearity function introduced in Eq. \ref{eq:CRlambda}. This metric employs the actual feature count rather than the nonlinearly transformed value used in Eq. \ref{eq:R1_nolineal}. Equation \ref{eq:R1} combines both objective scores into a single metric, enabling straightforward and objective comparison of configurations.
The function in Eq. \ref{eq:R1} identifies the most representative solution from the Pareto front by computing the Euclidean distance between each solution and the ideal point $(1,1)$ in objective space. The solution minimizing this distance maximizes $\hat{R_{1}}$. For specific applications, alternative functions can be employed to adjust the relative importance of objectives.
Once a representative solution is selected for each experiment through this criterion, statistical measures derived from these solutions quantify the impact of hyperparameters on algorithm performance.

%----------------------------
\subsubsection{Screening}
\label{sec:screening}

This phase identifies hyperparameter configurations that lead to good algorithm performance. Table \ref{tbl:screening_parametros} summarizes the analyzed hyperparameters and their explored values. These hyperparameters fall into two categories based on their computational implications. The first four parameters substantially affect both performance estimation across feature sets and computational cost. The remaining three parameters influence performance without introducing computational overhead.

%==============================================
\begin{table}[tb]
\centering
\caption{Summary of parameters considered in the screening.}
\begin{tabular}{ll}

\hline\hline

No. of fitness evaluations    & $\{ 1,3 \}$                            \\
No. of Objectives        & $\{ 2,3 \}$                                 \\
No. of Subordinate pops.     & $\{ 0,3 \}$                             \\
Replacement strategy & $\{ 'None', 'Parent', 'Selection' \}$           \\
Objective-II Sigmoid & $\{ True, False \}$                             \\
$\sigma$  & $\{ 0.001, 0.0025, 0.005, 0.015, 0.04, 0.08, 0.15, 0.3 \}$ \\
$\lambda$  & $\{ None, 0.1, 0.5, 1.5, 5, 15 \}$ \\ \hline\hline

\end{tabular}
\label{tbl:screening_parametros}
\end{table}
%==============================================

\begin{figure}[!tb]
\centering
    \includegraphics[trim = 0mm 0mm 0mm 0mm, clip, width=\textwidth]{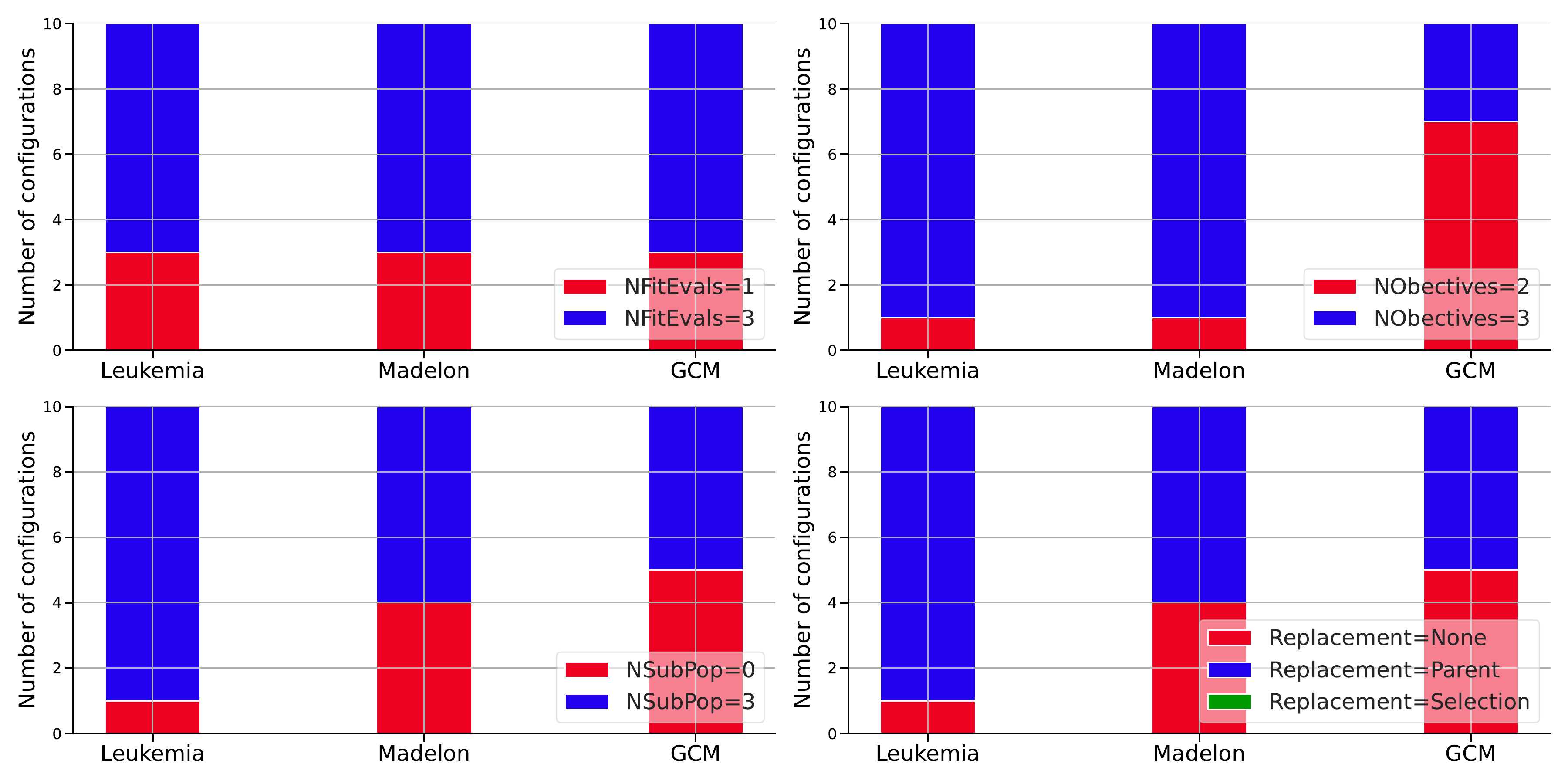} \\
    \caption{Frequency distributions for different hyperparameter settings considered in screening phase 1. Upper left: number of validation tests (1 test vs. 3 tests); Upper right: number of objectives (2 objectives vs. 3 objectives); Lower left: number of subordinate populations (SP) (no SP vs. 3 SP); Lower right: replacement strategy (PR, CR and SR).}
    \label{fig:screening1}
\end{figure}

\begin{figure}[!tb]
\centering
    \includegraphics[trim = 0mm 3cm 0mm 0mm, clip, width=\textwidth]{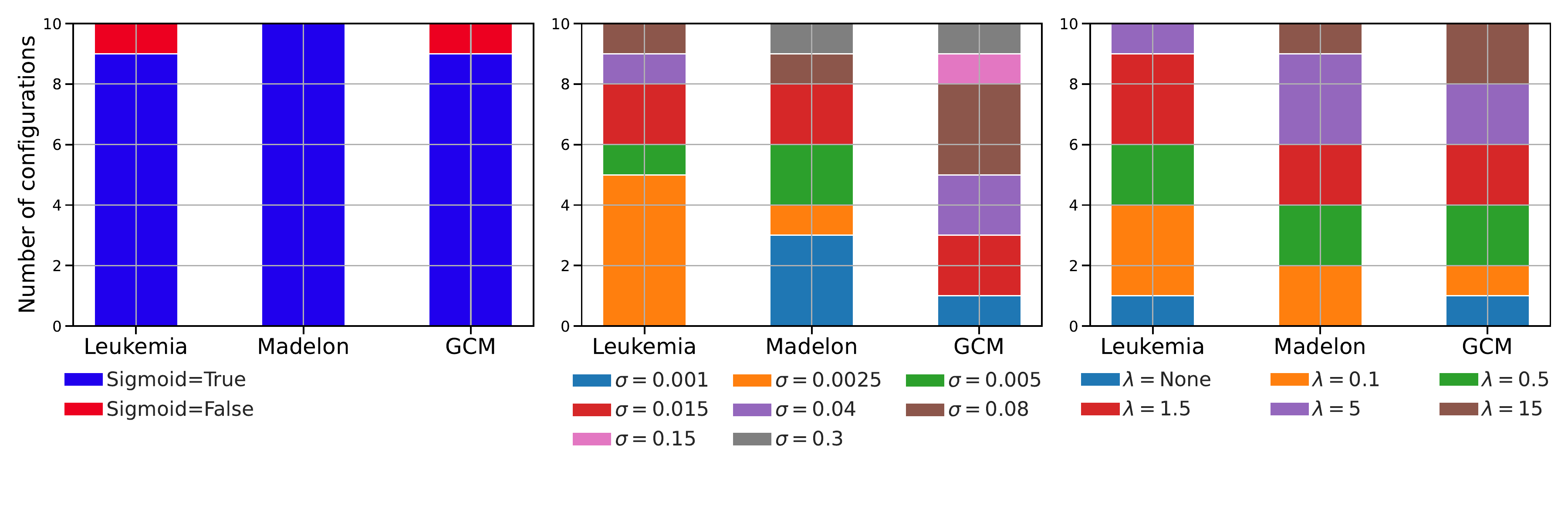} 
    \caption{Frequency distributions for different hyperparameter settings considered in screening phase 2. Left: use of the sigmoid function to normalize feature counts in objective function II; Center: Parameter $\sigma$ of the sharing function; Right: value of parameter $\lambda$ in sigmoid function.}
    \label{fig:screening2}
\end{figure}

To analyze the effect of hyperparameters on MOELIGA performance, experiments were processed as follows. For each replication, the best solution according to $\hat{R_{1}}$ was selected using Eq. \ref{eq:R1}, and the median $R_{1}$ value across replications characterized each configuration. The median was chosen over the mean due to the limited number of replications, necessitated by computational constraints, and its robustness to outliers. Configurations were then ranked in decreasing order of median $R_{1}$, and the top $10$ were retained for analysis. For these configurations, we examined the frequency distribution of each hyperparameter setting. Results are summarized in Figures \ref{fig:screening1} and \ref{fig:screening2}. 

%==========================================================================================================================
%==========================================================================================================================
Figure \ref{fig:screening1} shows results for the four hyperparameters affecting both performance and computational cost. Figures 
\ref{fig:screening1}a and \ref{fig:screening1}b demonstrate that using three test folds and incorporating the third objective function 
improve performance. Notably, the third objective appears in most top-ranked configurations despite not being included in the $R_{1}$ 
metric. For GCM, although only three configurations employed three objectives, these ranked within the top four by $R_{1}$ value.
Lower left and lower right panels of Figure \ref{fig:screening1} are identical, revealing a strong coupling between subordinate populations and the parent replacement strategy. Furthermore, configurations with three subordinate populations consistently demonstrate improved performance across experiments. 

%==========================================================================================================================
%==========================================================================================================================

Figure \ref{fig:screening2} yields two important conclusions. First, 
the left panel demonstrates that the sigmoid function (Eq. \ref{eq:CRlambda}) strongly correlates with performance improvement across 
all datasets, indicating that nonlinear mapping of feature count promotes solutions with higher $R_{1}$ values. Second,
center and right panels of Figure \ref{fig:screening2}
reveal that there is no clear association between specific sharing and sigmoid parameter ($\sigma$ and $\lambda$) values and optimal performance. All explored values for these hyperparameters appear in at least one top-ranked configuration across most datasets. Additionally, no correlation between $\lambda$ and $\sigma$ was observed.

In summary, this screening phase demonstrates that using three test folds, three objectives, subordinate populations, and the sigmoid function for the second objective collectively improve MOELIGA performance.

%---------------------------
\subsubsection{Refinement}

This refinement phase explores the effect of $\sigma$ and $\lambda$ on MOELIGA performance in greater detail. Given the reduced hyperparameter space, $10$ datasets were employed for this analysis. For $\lambda$, the explored values were $\{None,0.5,1.5,\\15\}$, where $None$ indicates that Eq. \ref{eq:CRlambda} is not applied and Eq. \ref{eq:featureratio} is used instead for the second objective. For $\sigma$, screening results showed that top-performing configurations predominantly used values within the range of $\left[ 0.001, 0.04 \right]$. Consequently, values $\{0.0025, 0.005, 0.025\}$ were selected, with the first two covering the lower range and the third representing higher values. Small $\sigma$ values enforce stricter niching, limiting the number of solutions per niche and promoting distribution across multiple niches \cite{deb2007}. This yields a more diverse Pareto-optimal set, enabling the selection of the most appropriate trade-off for specific applications.

Table \ref{tbl:refinement} summarizes the median $R_{1}$ values for each configuration. Values are color-coded within each dataset, with dark green indicating higher performance, dark red indicating lower performance, and lighter tones representing intermediate values. Row-wise analysis reveals limited variability within datasets, typically ranging between $0.05$ and $0.10$, indicating comparable performance across configurations. Leukemia exhibits slightly larger dispersion than other datasets.

Column-wise analysis across the ten datasets identifies three top-performing configurations: ($\lambda=1.5$, $\sigma=0.005$), ($\lambda=0.5$, $\sigma=0.0025$), and ($\lambda=1.5$, $\sigma=0.025$), achieving median $R_{1}$ values of $0.815$, $0.814$, and $0.812$, respectively. The corresponding standard deviations, shown in the last row of Table \ref{tbl:refinement}, are $0.105$, $0.094$, and $0.106$.
Considering both central tendency and dispersion, the configuration ($\lambda=0.5$, $\sigma=0.0025$) provides the optimal balance, exhibiting the lowest standard deviation among top performers. This is visually confirmed by its column containing the highest proportion of green and near-white cells.

%%%%%%%%%%%%%%%%%%%%%%%%%%%%%%%%%%%%%%%%%%%%%%%%%%%%%%%%%%%%%%%%%%%%%%%%%%%%%%%%%%%%%%%%%%%%%%%%%%%%%%%%%%%%%%%%%%%%%%%%%%%

\begin{table}[t]
\caption{Experimental results for the refinement stage in terms of $R_{1}$ metric. The $R_{1}$ values on each row are normalized to the 
range $\left[0,1\right]$, being $1.0$ the highest relative performance obtained from different experimental configurations. Values are 
colored according to their relative value.}
\label{tbl:refinement}
\scalebox{0.61}{\begin{tabular}{lcccccccccccc}
\hline\hline
\textbf{\Large{$\lambda$}}                                                 & \multicolumn{3}{c}{\textbf{None}}                                                                 & \multicolumn{3}{c}{\textbf{0.5}}                                                                                     & \multicolumn{3}{c}{\textbf{1.5}}                                                                  & \multicolumn{3}{c}{\textbf{15}}                                                                   \\ \hline
\textbf{\Large{$\sigma$}}                                                  & \textbf{0.0025} & \textbf{0.005} & \textbf{0.025} & \textbf{0.0025}                    & \textbf{0.005} & \textbf{0.025} & \textbf{0.0025} & \textbf{0.005} & \textbf{0.025} & \textbf{0.0025} & \textbf{0.005} & \textbf{0.025} \\ \hline
{\color[HTML]{000000} \textbf{dermatology}}  & \cellcolor[HTML]{57BB8A}0.850           & \cellcolor[HTML]{57BB8A}0.850          & 0.849          & \cellcolor[HTML]{F6D4D1}0.820                              & \cellcolor[HTML]{57BB8A}0.850          & \cellcolor[HTML]{CDEBDD}0.849          & 0.849           & \cellcolor[HTML]{FEFCFC}0.847          & 0.849          & \cellcolor[HTML]{E67C73}0.760           & \cellcolor[HTML]{EEAAA4}0.791          & \cellcolor[HTML]{EEAAA4}0.791          \\
{\color[HTML]{000000} \textbf{movement}}     & \cellcolor[HTML]{DDF2E8}0.787           & \cellcolor[HTML]{DAF0E5}0.787          & \cellcolor[HTML]{93D4B4}0.800          & \cellcolor[HTML]{FCF4F3}0.778                              & \cellcolor[HTML]{FBEEED}0.777          & \cellcolor[HTML]{8DD1B0}0.801          & \cellcolor[HTML]{57BB8A}0.811           & \cellcolor[HTML]{F4FBF8}0.782          & \cellcolor[HTML]{F7D8D5}0.773          & \cellcolor[HTML]{F7D9D6}0.773           & \cellcolor[HTML]{E67C73}0.755          & \cellcolor[HTML]{F8DDDB}0.774          \\
{\color[HTML]{000000} \textbf{arrhythmia}}   & \cellcolor[HTML]{6CC499}0.661           & \cellcolor[HTML]{5CBD8D}0.662          & \cellcolor[HTML]{F8DCDA}0.636          & \cellcolor[HTML]{57BB8A}0.663                              & \cellcolor[HTML]{F9E2E0}0.637          & \cellcolor[HTML]{FDF6F6}0.642          & \cellcolor[HTML]{E67C73}0.613           & \cellcolor[HTML]{EDF8F3}0.646          & \cellcolor[HTML]{F5CCC9}0.632          & \cellcolor[HTML]{FAE9E7}0.638           & \cellcolor[HTML]{D5EEE2}0.649          & \cellcolor[HTML]{EFF9F4}0.646          \\
{\color[HTML]{000000} \textbf{madelon}}      & \cellcolor[HTML]{FDF5F5}0.778           & \cellcolor[HTML]{E67C73}0.765          & \cellcolor[HTML]{8FD2B1}0.790          & \cellcolor[HTML]{57BB8A}0.795                              & \cellcolor[HTML]{FEFBFA}0.779          & \cellcolor[HTML]{F0B1AB}0.771          & \cellcolor[HTML]{FBFEFD}0.779           & \cellcolor[HTML]{E68077}0.766          & \cellcolor[HTML]{FEFBFA}0.779          & \cellcolor[HTML]{F3FAF7}0.780           & \cellcolor[HTML]{61BF91}0.794          & \cellcolor[HTML]{DEF2E8}0.782          \\
{\color[HTML]{000000} \textbf{isolet}}       & \cellcolor[HTML]{A2DABE}0.868           & \cellcolor[HTML]{57BB8A}0.872          & \cellcolor[HTML]{A9DCC3}0.867          & \cellcolor[HTML]{FEFDFD}0.862                              & \cellcolor[HTML]{FCF0EF}0.861          & \cellcolor[HTML]{7CCAA4}0.870          & \cellcolor[HTML]{FEFFFE}0.863           & \cellcolor[HTML]{F6D3D0}0.858          & \cellcolor[HTML]{D2EDE0}0.865          & \cellcolor[HTML]{FDF8F8}0.862           & \cellcolor[HTML]{FCF3F2}0.861          & \cellcolor[HTML]{E67C73}0.849          \\
{\color[HTML]{000000} \textbf{mfeat}}        & \cellcolor[HTML]{C3E7D6}0.953           & \cellcolor[HTML]{FEFBFA}0.951          & \cellcolor[HTML]{96D5B6}0.955          & \cellcolor[HTML]{BCE4D0}0.954                              & \cellcolor[HTML]{FEFAFA}0.951          & \cellcolor[HTML]{F8DFDD}0.948          & \cellcolor[HTML]{A1D9BE}0.955           & \cellcolor[HTML]{57BB8A}0.958          & \cellcolor[HTML]{F6FCF9}0.951          & \cellcolor[HTML]{F3C5C1}0.945           & \cellcolor[HTML]{F5CBC7}0.946          & \cellcolor[HTML]{E67C73}0.939          \\
{\color[HTML]{000000} \textbf{leukemia}}     & \cellcolor[HTML]{E78279}0.700           & \cellcolor[HTML]{E67C73}0.689          & \cellcolor[HTML]{E67C73}0.689          & \cellcolor[HTML]{85CEAA}0.914                              & \cellcolor[HTML]{9ED8BB}0.914          & \cellcolor[HTML]{F6FBF9}0.914          & \cellcolor[HTML]{FEFEFE}0.914           & \cellcolor[HTML]{57BB8A}0.914          & \cellcolor[HTML]{FEFEFE}0.914          & \cellcolor[HTML]{99D6B8}0.914           & \cellcolor[HTML]{99D6B8}0.914          & \cellcolor[HTML]{FEFEFE}0.913          \\
{\color[HTML]{000000} \textbf{all-leukemia}} & \cellcolor[HTML]{E67C73}0.789           & \cellcolor[HTML]{ACDEC6}0.829          & \cellcolor[HTML]{EFAFA9}0.798          & \cellcolor[HTML]{C2E6D4}0.825                              & \cellcolor[HTML]{F9FDFB}0.813          & \cellcolor[HTML]{EDA49E}0.796          & \cellcolor[HTML]{EDA49E}0.796           & \cellcolor[HTML]{FDF7F7}0.811          & \cellcolor[HTML]{BAE3CF}0.826          & \cellcolor[HTML]{E68077}0.790           & \cellcolor[HTML]{57BB8A}0.846          & \cellcolor[HTML]{F3FBF7}0.815          \\
{\color[HTML]{000000} \textbf{yeoh}}         & \cellcolor[HTML]{FAE4E3}0.870           & \cellcolor[HTML]{F4C9C5}0.865          & \cellcolor[HTML]{E6F5EE}0.880          & \cellcolor[HTML]{F5CCC8}0.865                              & \cellcolor[HTML]{F1B8B3}0.862          & \cellcolor[HTML]{59BC8B}0.908          & \cellcolor[HTML]{E67C73}0.850           & \cellcolor[HTML]{57BB8A}0.908          & \cellcolor[HTML]{CFECDD}0.885          & \cellcolor[HTML]{A6DBC1}0.893           & \cellcolor[HTML]{EEA6A0}0.858          & \cellcolor[HTML]{94D4B4}0.896          \\
{\color[HTML]{000000} \textbf{gcm}}          & \cellcolor[HTML]{FEFDFC}0.638           & \cellcolor[HTML]{B1E0C9}0.649          & \cellcolor[HTML]{FCF4F3}0.634          & \cellcolor[HTML]{57BB8A}0.661                              & \cellcolor[HTML]{EEA9A3}0.598          & \cellcolor[HTML]{E67C73}0.576          & \cellcolor[HTML]{FEFEFE}0.639           & \cellcolor[HTML]{79C9A2}0.657          & \cellcolor[HTML]{BAE4CF}0.648          & \cellcolor[HTML]{FCEFEE}0.631           & 0.639          & \cellcolor[HTML]{FCFEFD}0.639          \\ \hline
\multicolumn{1}{l}{\textbf{avg.}}                                & \multicolumn{1}{l}{0,789}               & \multicolumn{1}{l}{0,792}              & \multicolumn{1}{l}{0,790}              & \multicolumn{1}{l}{\textbf{0,814}} & \multicolumn{1}{l}{0,804}              & \multicolumn{1}{l}{0,807}              & \multicolumn{1}{l}{0,807}               & \multicolumn{1}{l}{\textbf{0,815}}              & \multicolumn{1}{l}{\textbf{0,812}}              & \multicolumn{1}{l}{0,799}               & \multicolumn{1}{l}{0,805}              & \multicolumn{1}{l}{0,804}              \\
\multicolumn{1}{l}{\textbf{std.}}                            & \multicolumn{1}{l}{0.102}               & \multicolumn{1}{l}{0.100}              & \multicolumn{1}{l}{0.107}              & \multicolumn{1}{l}{\textbf{0.094}} & \multicolumn{1}{l}{0.113}              & \multicolumn{1}{l}{0.120}              & \multicolumn{1}{l}{0.109}               & \multicolumn{1}{l}{0.105}              & \multicolumn{1}{l}{0.106}              & \multicolumn{1}{l}{0.107}               & \multicolumn{1}{l}{0.102}              & \multicolumn{1}{l}{0.102}              \\ \hline\hline
\end{tabular}}
\end{table}
%%%%%%%%%%%%%%%%%%%%%%%%%%%%%%%%%%%%%%%%%%%%%%%%%%%%%%%%%%%%%%%%%%%%%%%%%%%%%%%%%%%%%%%%%%%%%%%%%%%%%%%%%%%%%%%%%%%%%%%%%%%

\subsection{Pareto Front Analysis}
\label{subsec:pareto}

This section analyzes the behavior of Pareto fronts generated by MOELIGA from initialization (Generation 1) through final evolution (Generation 300). We selected three representative datasets spanning different dimensionality ranges: \emph{Movement} (90 features), \emph{Mfeat} (649 features), and \emph{GCM} (16,063 features). In all figures below, small orange dots represent the initial Pareto front, while large colored dots represent the final front, with colors ranging from dark tones to bright yellow. The color scale represents the $\hat{R}_{1}$ metric (Eq. \ref{eq:R1}), which measures solution quality as the Euclidean distance to the ideal point in normalized objective space, where brighter colors indicate superior trade-offs between classification performance and feature dimensionality.

\begin{figure}[!tb]
\centering
    \includegraphics[trim = 0mm 0mm 0mm 0mm, clip, width=\textwidth]{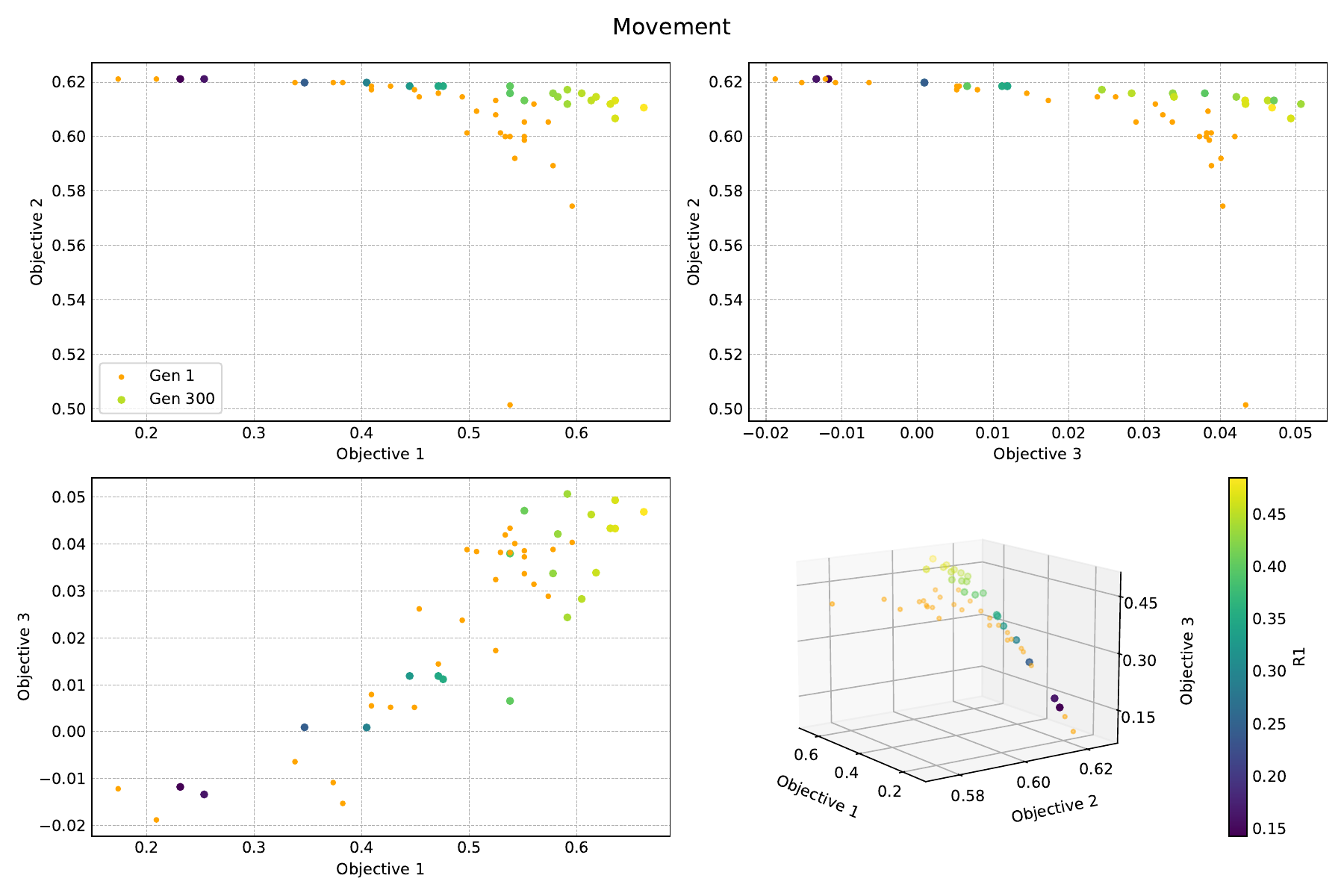}
    \caption{Comparison of the Pareto Fronts in the first and last generation for dataset Movement. Small orange dots correspond to the Pareto solutions from the first generation, while large dots represent the solutions from the last generation and are colored by the corresponding values for R1.}
    \label{fig:pareto_movement}
\end{figure}

Figure \ref{fig:pareto_movement} presents the \emph{Movement} dataset evolution across four views: Objective 1 (UAR) versus Objective 2 (cardinality ratio) in the top-left panel, revealing the accuracy-dimensionality trade-off; Objective 2 versus Objective 3 (geometry-based metric $M_{d}$) in the top-right; Objective 1 versus Objective 3 in the bottom-left, showing their correlation; and a three-dimensional integration in the bottom-right. The scattered orange dots represent the initial random population, while the large colored dots correspond to the last generation and form a structured front, where bright yellow tones identify optimal trade-off regions.

The final front in the left-top panel of Figure \ref{fig:pareto_movement} exhibits a concave shape spanning UAR from $0.25$ to $\approx 0.65$ with around $35$ features. The brightest yellow solution lies at approximately $35$ features and $UAR \approx 0.66$, representing an optimal balance. Comparing generations, median UAR improved from $0.516$ to $0.544$ ($+5.6\%$), while median features decreased slightly from $35$ to $34.5$ ($-1.5\%$). Evolution occurs predominantly along the UAR axis, evidenced by the horizontal displacement of colored dots above orange dots at similar vertical positions.

Three algorithmic factors explain this pattern. First, UAR requires computationally intensive classifier training across multiple splits (Section \ref{proposal:uar}), making it informationally rich compared to trivially-computed feature counts, since the evolutionary local improvement strategy (Section \ref{proposal:evolutionary_local_improvement}) specifically intensifies search along this objective. Second, staggered initialization (Section \ref{sec:init}) pre-establishes feature diversity by initializing population segments with $3\%$, $15\%$, and $35\%$ active genes, producing solutions spanning $3$ to $32$ features, whereas all initial solutions cluster at low UAR ($0.25-0.35$), concentrating selective pressure on Objective 1. Third, the sigmoid transformation (Equation \ref{eq:CRlambda}) guides solutions toward compact subsets, after which further reduction conflicts with accuracy, limiting evolution along this axis.

The final front in the left-bottom panel of Figure \ref{fig:pareto_movement} spans $M_d$ values (Objective 3) from $-0.015$ to $0.051$ 
(median $0.04$). These near-zero values are expected: $M_d$ measures class separability with bounds $[-1, 1]$, where $M_d = 1$ 
indicates perfect linear separability and $M_d = 0$ represents balanced intra/inter-class distances. For real-world multi-class 
problems like \emph{Movement}'s $15$-class gesture recognition, perfect separability is unattainable due to inherent class overlap. 
Objective 3 functions as a regularization mechanism promoting classifier-independent features rather than a maximization target. 

\begin{figure}[!tb]
\centering
    \includegraphics[trim = 0mm 0mm 0mm 0mm, clip, width=\textwidth]{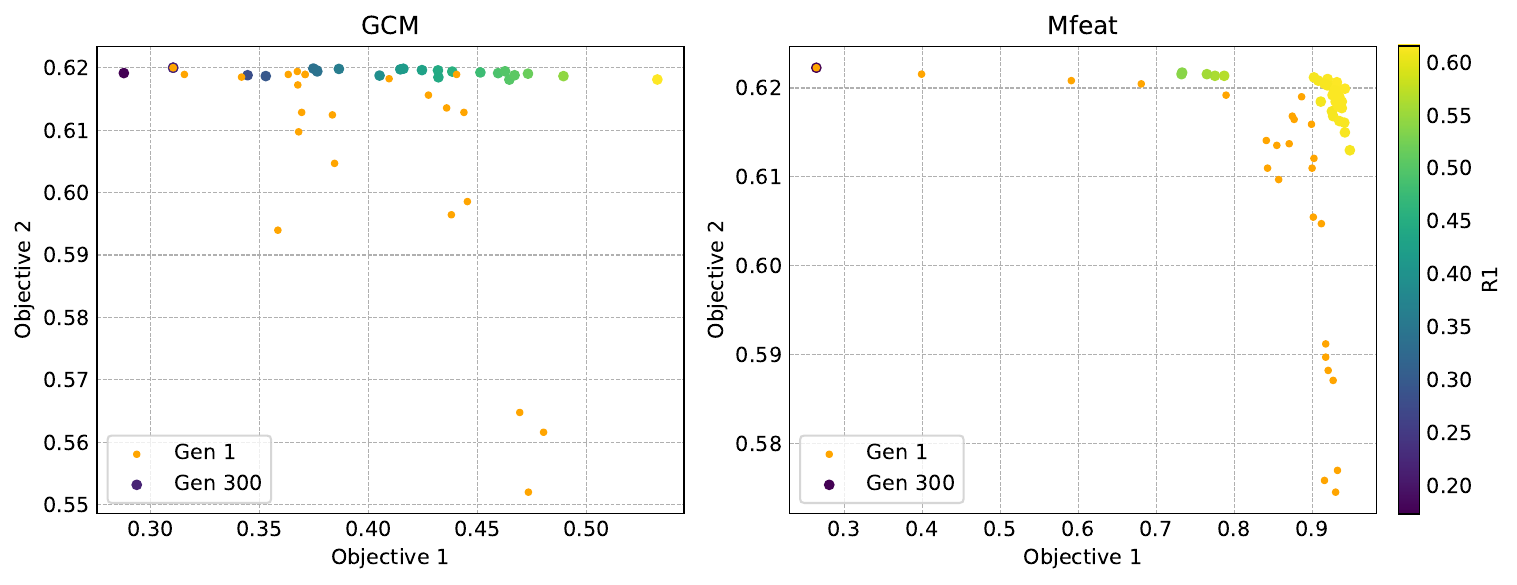}
    \caption{Comparison of the Pareto Fronts in the first and last generation for datasets GCM (left) and Mfeat (right), considering the pair of most relevant objectives (Objective 1 and Objective 2). Small orange dots correspond to the Pareto solutions from the first generation, while large dots represent the solutions from the last generation and are colored by the corresponding values for R1.}
    \label{fig:pareto_gcm_mfeat}
\end{figure}

Figure \ref{fig:pareto_gcm_mfeat} (right panel) shows \emph{Mfeat}'s evolution, demonstrating consistent patterns: median UAR improved from $0.887$ to $0.928$ ($+4.6\%$) with features decreasing from $252$ to $247$ ($-2.0\%$). Unlike \emph{Movement}'s sharp concave front, \emph{Mfeat} exhibits a gentler trade-off with multiple bright yellow solutions ($\hat{R}_{1} \approx 0.55-0.60$) distributed across the feature range, indicating a plateau region where feature count can vary without severe performance degradation. Solutions with $14$, $16$, $18$, and $20$ features achieve UAR within $0.02$ of each other ($0.93-0.95$), providing practitioners flexibility in selecting feature budgets while maintaining near-optimal trade-offs. The optimal solution identified by $\hat{R}_{1}$ achieves $UAR \approx 0.95$ with approximately $251$ features.

Figure \ref{fig:pareto_gcm_mfeat} (left panel) presents \emph{GCM}, where the final front exhibits near-vertical structure spanning UAR from $0.36$ to $0.52$ with $5900$ to $6400$ features (approximately $37-40\%$ of the total $16,063$ features). Median UAR increased from $0.377$ to $0.420$ ($+11.4\%$), while features decreased marginally from $6191$ to $6115$ ($-1.2\%$). Bright yellow solutions (optimal $\hat{R}_{1} \approx 0.35-0.40$) concentrate around $6134$ features with $UAR \approx 0.53$. This vertical pattern reflects high-dimensional genomic data characteristics: most features are non-informative, and once a core discriminative set is identified ($\sim6100$ features), performance gains require refining composition rather than expanding size. The color gradient from dark tones at UAR $0.36$ to bright yellow at $0.52$ within this narrow feature window visualizes how substituting features substantially impacts classification.

Across all three dimensionality regimes, MOELIGA demonstrates consistent evolution with UAR improvements of $5.6\%$, $4.6\%$, and $11.4\%$ for \emph{Movement}, \emph{Mfeat}, and \emph{GCM} respectively, coupled with feature reductions of $1.5\%$, $2.0\%$, and $1.2\%$. Final fronts contain $18$--$32$ non-dominated solutions spanning $>0.20$ UAR units, ensuring diverse trade-offs. Bright yellow solutions (optimal $\hat{R}_{1}$ values) concentrate at $35$ features/UAR $0.66$ (\emph{Movement}), $251$ features/UAR $0.95$ (\emph{Mfeat}), and $6134$ features/UAR $0.53$ (\emph{GCM}). Evolution predominantly along the UAR axis reflects algorithmic design where evolutionary local improvement targets classification performance, staggered initialization pre-establishes feature diversity, and sigmoid transformation guides compactness, while Objective 3 serves its regularization role as confirmed by near-zero values arising from inherent class overlap.

%===========================================
\subsection{Comparison with SOTA approaches}
\label{subsec:comparison_SOTA}

In order to compare the performance of MOELIGA we have selected different types of feature selection methods, based in recent and relevant literature, including wrappers and filter approaches:

\begin{itemize}
    \item  \textbf{Recursive Feature Elimination} (RFE) starts with the complete set of features and recursively eliminates the least useful feature, decreasing the size of the selected set in one at each step \cite{guyon2002}. 
    In the same way we used a DT classifier as objective function in our approach, we used a DT classifier as estimator for RFE using 
scikit-learn library for python \cite{scikit}.
    % KBEST + F-Classif
    \item \textbf{ANOVA F-value} (Kbest) method consists in selecting the $K$ top ranked features according to the ANOVA F-value \cite{elssied2014}. Here we used the \emph{K-best} ranking method and \emph{F-Classif}, both from scikit-learn \cite{scikit}.
    \item \textbf{Mutual Information} (MI) ranking is an univariate feature selection approach, which ranks the features according to the dependence between each of them and the output. The scikit-learn \cite{scikit} implementation relies on non-parametric methods based on entropy estimation from $k$-nearest neighbors distances \cite{kraskov2004,ross2014}.
    \item \textbf{Sequential Forward Selection} (SFS) incorporates features to compose a subset in a greedy fashion \cite{pudil1994}. It starts with an empty subset and recursively incorporates one feature in each step, choosing the feature that maximizes the score of the current subset. 
    \item \textbf{Sequential Forward Floating Selection} (SFFS) incorporates certain flexibility to allow reconsidering characteristics 
that had been previously incorporated in the subset, aiming to avoid the nesting effect \cite{ververidis2008}. For both SFS and SFFS 
methods we used the python implementation provided by \emph{mlxtend} \cite{raschkas2018} with DT classifier as estimator.
    \item  \textbf{Boruta} \cite{kursa2010a} is a feature selection method based on  Random Forest (RF) \cite{breiman2001}. 
    It iterates comparing the importance of the original features with the importance of randomised copies. Then, only the features of 
higher importance than that of the randomised variables are considered relevant \cite{kursa2010a}. In our experiments we used the 
python implementation provided in the Boruta package \cite{kursa2010b}, with the number of trees  automatically determined. 
    \item \textbf{ReliefF}~\cite{kononenko1996} is an improved version of Relief \cite{Urbanowicz2017} method, which  calculates a 
score for each feature based on  value difference between pairs of nearest neighbor instances. 
% ReliefF uses the Manhattan (L1) norm to account for the distance of neighboring instance pairs. 
    ReliefF searches for a number $k$ of nearest neighbors, instead of finding the single nearests hit and miss, averaging their 
contribution to the weights for each feature. This counteracts the negative effect of noisy features on selection of the nearest 
neighbors.
    \item \textbf{SURF} is another algorithm inspired on Relief, which uses a distance threshold $T$ for determining which instances are to be considered neighbors \cite{greene2009}. In this manner, SURF eliminates the user-defined parameter $k$. The threshold $T$ is determined by the average distance of all instance pairs in the training set.
    \item \textbf{SURF*} (or SURFstar) is an extension of SURF that incorporates the use of \emph{far} instances, with inverted scoring, as well near neighbors in weight updates \cite{greene2010}. The bibliography reports improved detection ability for 2-way epistatic interactions in gene expression data \cite{Urbanowicz2017}.
    \item \textbf{MultiSURF*} (MSurf*) is an extension of SURF* which adapts the near and far neighborhood boundaries according to the average distance and deviation from the target instance \cite{granizo2013}. Based on the standard deviation, MultiSURF* defines a middle-distance zone where instances do not contribute to scoring. This algorithm has been proven to perform better in the detection of pure 2-way feature interactions \cite{Urbanowicz2017}.
    \item  \textbf{MultiSURF} (MSurf) is a simplification of MultiSURF*  which preserves the middle-distance zone and the determination of neighborhood based on the distance to target instances, and it does not account for \emph{far} instances for updating weights \cite{Urbanowicz2017}. The authors report improved detection of 2-way and 3-way interactions and univariate associations. 
    For the all these Relief-based methods we have used the python implementations provided in  ReBATE\footnote{\url{https://epistasislab.github.io/scikit-rebate/}} \cite{Urbanowicz2017}.
\end{itemize}

The performance of MOELIGA was compared against the $11$ feature selection methods listed above, considering $14$ classification datasets (see Table \ref{tbl:datasets}). Since these methods require specifying the number of features to select, the same number as obtained with MOELIGA was employed for each dataset. UAR values obtained with each method are presented in Table \ref{tbl:SOTA_comparison}. Each result in the table corresponds to the median over $5$ repetitions for each combination of methods and datasets.

\begingroup
\setlength{\tabcolsep}{3pt} % Default value: 6pt
\begin{table}[t]
\caption{Comparison of MOELIGA with state-of-the-art methods in terms of the median UAR on test data across different datasets. The best results for each dataset are indicated in boldface and, for comparison, second-best results are underlined.}
\label{tbl:SOTA_comparison}
\scalebox{0.8}{\begin{tabular}{
>{\columncolor[HTML]{FFFFFF}}l 
>{\columncolor[HTML]{FFFFFF}}r
>{\columncolor[HTML]{FFFFFF}}c 
>{\columncolor[HTML]{FFFFFF}}c 
>{\columncolor[HTML]{FFFFFF}}c 
>{\columncolor[HTML]{FFFFFF}}c 
>{\columncolor[HTML]{FFFFFF}}c 
>{\columncolor[HTML]{FFFFFF}}c 
>{\columncolor[HTML]{FFFFFF}}c 
>{\columncolor[HTML]{FFFFFF}}c 
>{\columncolor[HTML]{FFFFFF}}c 
>{\columncolor[HTML]{FFFFFF}}c 
>{\columncolor[HTML]{FFFFFF}}c 
>{\columncolor[HTML]{FFFFFF}}c }
\hline
\hline
 &
  \rotatebox[origin=c]{60}{\textbf{Features}} &  
  \rotatebox[origin=c]{60}{\textbf{Moeliga}} &
  \rotatebox[origin=c]{60}{\textbf{ReliefF}} &
  \rotatebox[origin=c]{60}{\textbf{Surf}} &
  \rotatebox[origin=c]{60}{\textbf{Surf$\star$}} &
  \rotatebox[origin=c]{60}{\textbf{MSurf}} &
  \rotatebox[origin=c]{60}{\textbf{MSurf$\star$}} &
  \rotatebox[origin=c]{60}{\textbf{MI}} &
  \rotatebox[origin=c]{60}{\textbf{RFE}} &   % RFE-DT 
  \rotatebox[origin=c]{60}{\textbf{KBest}} & % KBest-Ftest
  \rotatebox[origin=c]{60}{\textbf{SFS}} &   % SFS-DT
  \rotatebox[origin=c]{60}{\textbf{SFFS}} &  % SFFS-DT
  \rotatebox[origin=c]{60}{\textbf{Boruta}} \\ \hline  % {RF (ne=auto)} 
{\color[HTML]{000000} \textbf{derma}} &
  6/34 & \textbf{0.96} & 0.70 & 0.69 & 0.54 & 0.65 & 0.53 & 0.67 & 0.68 & 0.50 & {\ul 0.89} & {\ul 0.89} & 0.72 \\
{\color[HTML]{000000} \textbf{optd}} &
  10/64 & \textbf{0.91} & 0.79 & 0.81 & 0.66 & 0.83 & 0.52 & 0.81 & 0.61 & 0.81 & {\ul 0.85} & {\ul 0.85} & 0.48 \\
{\color[HTML]{000000} \textbf{move}} &
  9/90 & \textbf{0.80} & 0.38 & 0.48 & 0.28 & 0.38 & 0.28 & 0.45 & 0.44 & 0.48 & {\ul 0.60} & {\ul 0.60} & 0.32 \\
{\color[HTML]{000000} \textbf{arrh}} &
  11/279 &\textbf{0.67} & 0.28 & 0.28 & 0.23 & 0.33 & 0.26 & 0.28 & 0.28 & 0.26 & {\ul 0.38} & 0.35 &  0.25 \\
{\color[HTML]{000000} \textbf{mad}} &
  22/500 & 0.80 & {\ul 0.83} & 0.83 & 0.82 &  0.82 & \textbf{0.84} & 0.63 & 0.64 & 0.76 & 0.77 & 0.77 &  0.83 \\
{\color[HTML]{000000} \textbf{smart}} &
  8/561 & \textbf{0.88} & 0.66 & 0.60 & 0.60 & 0.60 & 0.60 & 0.53 & 0.64 & 0.67 & {\ul 0.79} & {\ul 0.79} & 0.56 \\
{\color[HTML]{000000} \textbf{isolet}} &
  44/617 &\textbf{0.88} & 0.69 & 0.48 & 0.47 & 0.47 & 0.45 & 0.62 & 0.45 & 0.66 & 0.79 & {\ul 0.80} &  0.33 \\
{\color[HTML]{000000} \textbf{mfeat}} &
  14/649 & \textbf{0.96} & 0.91 & 0.84 & 0.57 & 0.74 & 0.74 & 0.90 & 0.74 & 0.91 & {\ul 0.92} & {\ul 0.92} &  0.75 \\
{\color[HTML]{000000} \textbf{gisette}} &
  189/$\sim$5k & 0.91 & 0.93 & 0.93 & \textbf{0.93} &  0.91 & {\ul 0.93} & 0.92 & 0.91 & 0.93 & 0.90 & 0.92 &  0.92 \\
{\color[HTML]{000000} \textbf{leuk}} &
  30/$\sim$7k & \textbf{0.91} & \textbf{0.91} & \textbf{0.91} & 0.65 & \textbf{0.91} & 0.76 & \textbf{0.91} & \textbf{0.91} &  \textbf{0.91} &  \textbf{0.91} & \textbf{0.91} & 0.65 \\
{\color[HTML]{000000} \textbf{all-leuk}} &
  168/$\sim$12k &\textbf{0.83} & 0.70 & 0.67 & 0.67 & 0.65 & 0.74 & 0.68 & 0.63 & 0.70 & 0.65 &  0.66 & {\ul 0.78} \\
{\color[HTML]{000000} \textbf{yeoh}} &
  270/$\sim$12k & {\ul 0.87} & 0.78 & 0.68 & 0.78 & 0.74 & 0.77 & 0.84 & 0.78 & 0.77 & 0.79 & 0.74 & \textbf{0.97} \\
{\color[HTML]{000000} \textbf{gcm}} &
  264/$\sim$16k & \textbf{0.66} & 0.29 & {\ul 0.48} & 0.25 & 0.30 & 0.21 & 0.30 & 0.43 & 0.35 &  0.41 &  0.43 & 0.45 \\
{\color[HTML]{000000} \textbf{tcga}} &
  302/$\sim$20k & \textbf{0.98} &  0.96 & {\ul 0.98} &  0.97 &  0.97 &  0.96 &  0.94 &  0.95 &  0.96 &  0.97 &  0.96 &  0.95 \\
  \hline
\end{tabular}}
\end{table}
\endgroup

From the analysis of Table \ref{tbl:SOTA_comparison}, it can be noted that MOELIGA achieves superior performances in $9$ out of the $14$ analyzed datasets, with the largest observed difference being $30$\% compared to other methods.
Only for the \texttt{Yeoh} dataset does MOELIGA exhibit a notably lower performance ($10$\%) when it is compared to the better method. However, MOELIGA performance remains at the top of the other methods. For the remaining datasets (madelon and gisette), the performance of MOELIGA is equivalent to that of other algorithms, obtaining similar or slightly lower performances ($\sim 3$\%). Furthermore, in these cases, the best performances are achieved by different algorithms. Remarkably, methods based on ``SF'' more frequently obtain the second-best performance for datasets with up to $1000$ features, with UAR values at least $4$\% lower than MOELIGA. However, this changes when increasing the number of features, as algorithms with better performances vary.

To verify whether the observed performance differences among algorithms are statistically significant, we applied the Friedman test for complete block designs \cite{Demsar2006} across the $12$ feature selection algorithms and $14$ datasets. The test yielded $\chi^{2}_{F} = 39.70$ ($df = 11$, $p < 0.001$), and the Iman-Davenport correction confirmed this result ($FF = 4.52$, $p < 0.001$) \cite{Derrac2011}, rejecting the null hypothesis of equal performance across the evaluated datasets.

\begin{figure}[!tb]
\centering
    \includegraphics[trim = 0mm 0mm 0mm 12mm, clip, width=\textwidth]{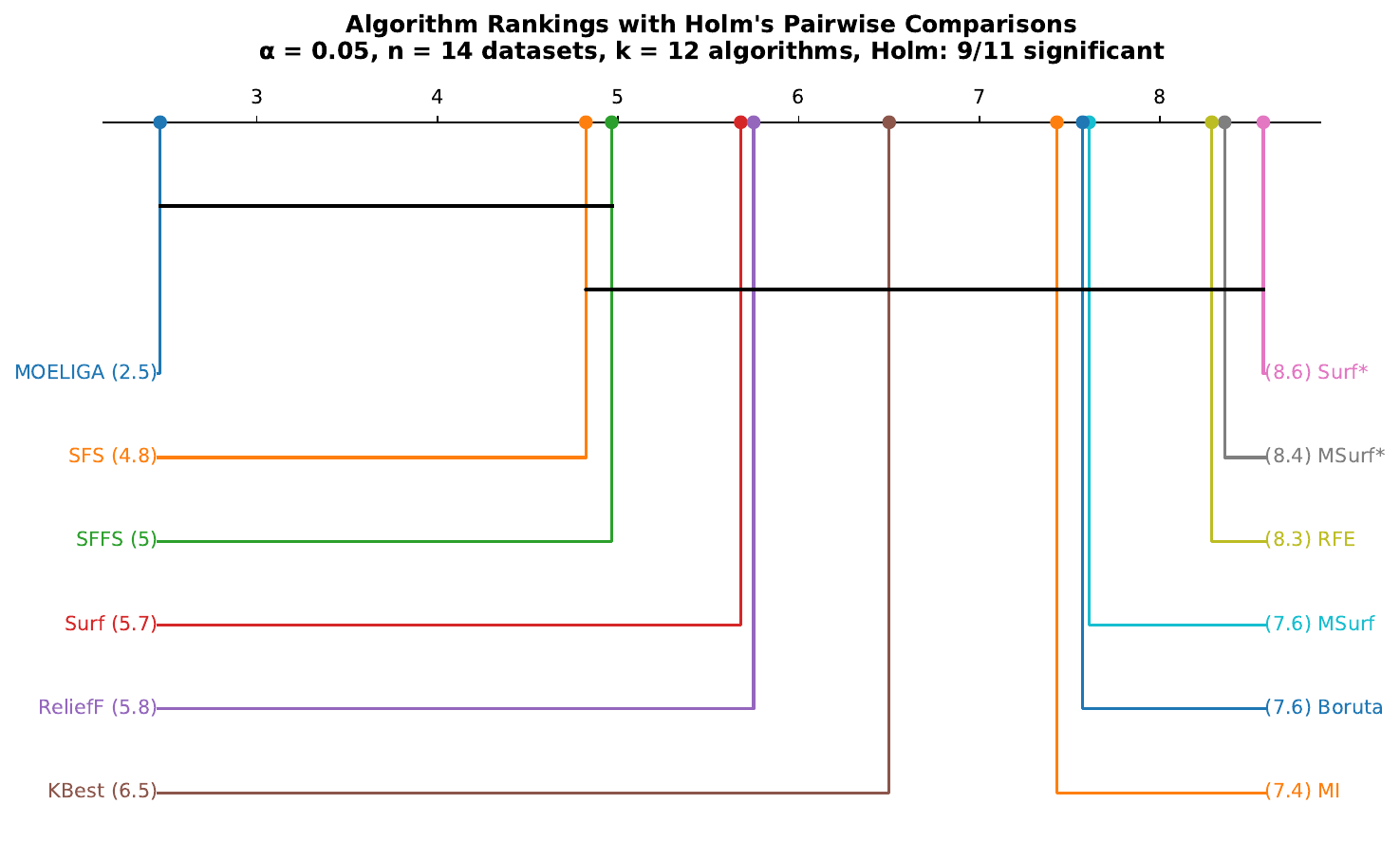} \\
    \caption{Algorithm rankings with pairwise significance testing using Holm's procedure ($\alpha = 0.05$, $n = 14$ datasets, $k = 12$ algorithms). Algorithms are ordered by average Friedman rank. Horizontal bars connect algorithms showing no significant difference from each other based on Holm's step-down correction.}
    \label{fig:cdiff_plot}
\end{figure}

To identify which specific algorithms differ from MOELIGA, we conducted post-hoc pairwise comparisons using the $z$-statistic derived from Friedman average ranks, with Holm's step-down procedure applied to control the family-wise error rate at $\alpha = 0.05$ \cite{Derrac2011}. MOELIGA achieved the best average rank ($R_{\text{avg}} = 2.46$) and demonstrated statistically significant improvements over $9$ of the $11$ competing algorithms after Holm's correction (adjusted $p < 0.05$); the differences with SFFS and SFS did not reach statistical significance. Figure~\ref{fig:cdiff_plot} illustrates these results through a Holm-adjusted ranking diagram, adopting the graphical convention of Critical Difference diagrams \cite{Demsar2006}, where the clique bar connecting MOELIGA exclusively to SFS and SFFS visually confirms the two non-significant pairwise comparisons, while all remaining $9$ competitors are unconnected to MOELIGA.

\begin{figure}[!tb]
\centering
    \includegraphics[trim = 10mm 0mm 20mm 10mm, clip, width=\textwidth]{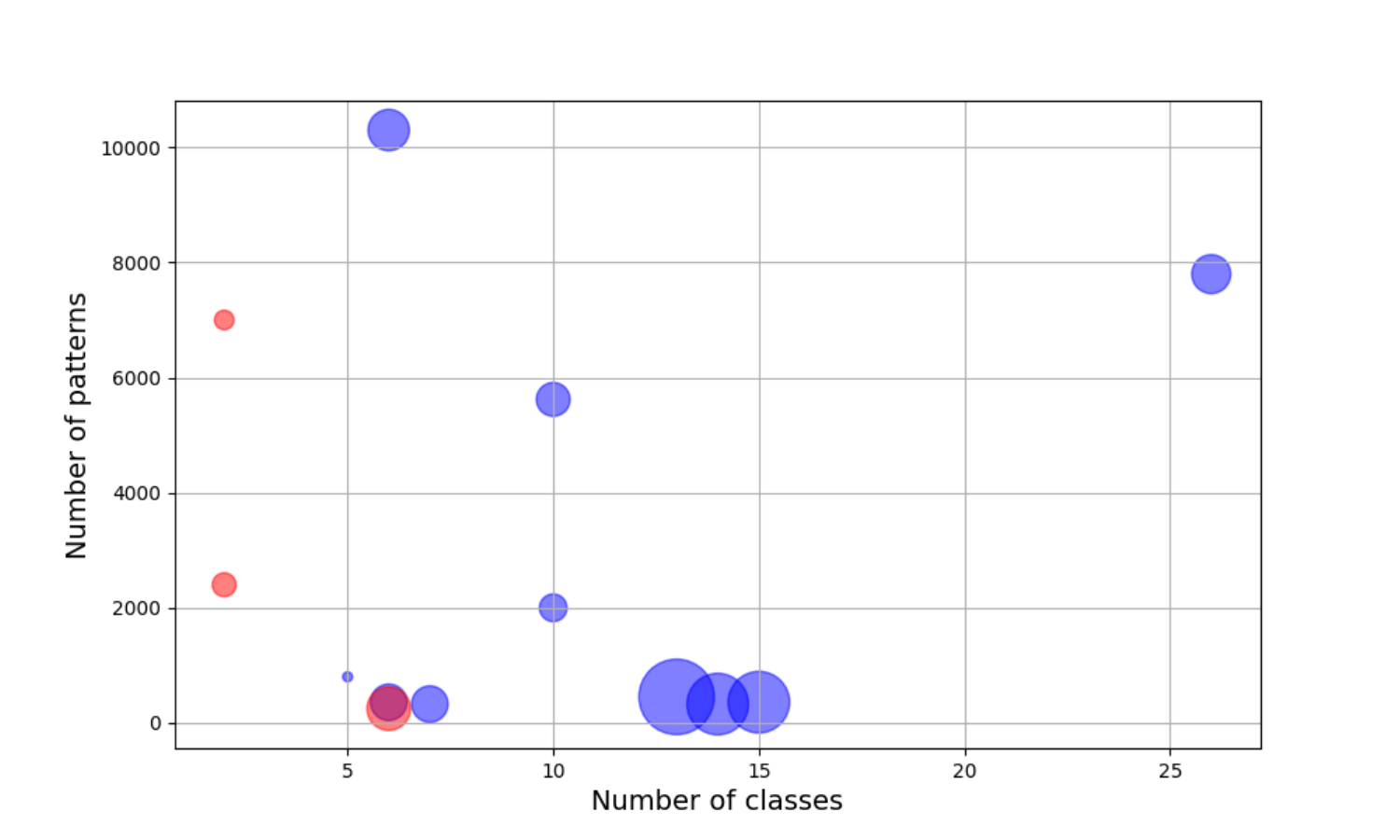} \\
    \caption{UAR difference between MOELIGA and the best-performing competitors for each dataset. Circle size denotes the magnitude of performance difference; blue circles indicate differences in favor of MOELIGA and red circles indicate the opposite.}
    \label{fig:MOELIGAvsSOTA}
\end{figure}

To understand the conditions under which MOELIGA attains higher performance, a chart showing the difference in UAR between MOELIGA and the best-performing competing method was generated. The result is displayed in Figure \ref{fig:MOELIGAvsSOTA}, which considers 
the best  competitors from Table \ref{tbl:SOTA_comparison} for each dataset.
Each point represents a dataset characterized by its number of classes and number of patterns, with the performance difference indicated by both the color and diameter of the circles. Larger circles denote greater performance differences, with blue circles indicating cases where MOELIGA outperforms the competitor and red circles indicating the opposite. As observed, MOELIGA achieves better performance as the number of classes increases or when a large number of patterns is available. When both conditions are met simultaneously, MOELIGA demonstrates a clear and substantial advantage.

%==========================================================================================================================
%==========================================================================================================================

%==========================================================================================================================
%==========================================================================================================================

Finally, an important characteristic of MOELIGA is its ability to automatically determine the number of features during the optimization process. In contrast, competing methods require this parameter to be specified \textit{a priori}, which can lead to suboptimal results when the predefined value does not match the intrinsic dimensionality of the problem. The automatic adaptation of feature cardinality is particularly relevant, as an inadequate number of features --either excessive or insufficient-- directly impacts the generalization capability of subsequent classification models.

%%%%%%%%%%%%%%%%%%%%%%%%%%%%%%%%%%%%%%%%%%%%%%%%%%%%%%%%%%%%%%%%%%%%%%%%%%%%%%%%%%%%%%%%%%%%%%%%%%%%%%%%%%%%%%%%%%%%%%%%%%%

\subsection{Comparing MOELIGA with MI and SFFS in searching the optimal feature subset size}
\label{sec:barrido_sota}

%%%%%%%%%%%%%%%%%%%%%%%%%%%%%%%%%%%%%%%%%%%%%%%%%%%%%%%%%%%%%%%%%%%%%%%%%%%%%%%%
\begin{figure}[!tb]
\centering
    \includegraphics[trim = 0mm 5mm 0mm 0mm, clip, width=\textwidth]{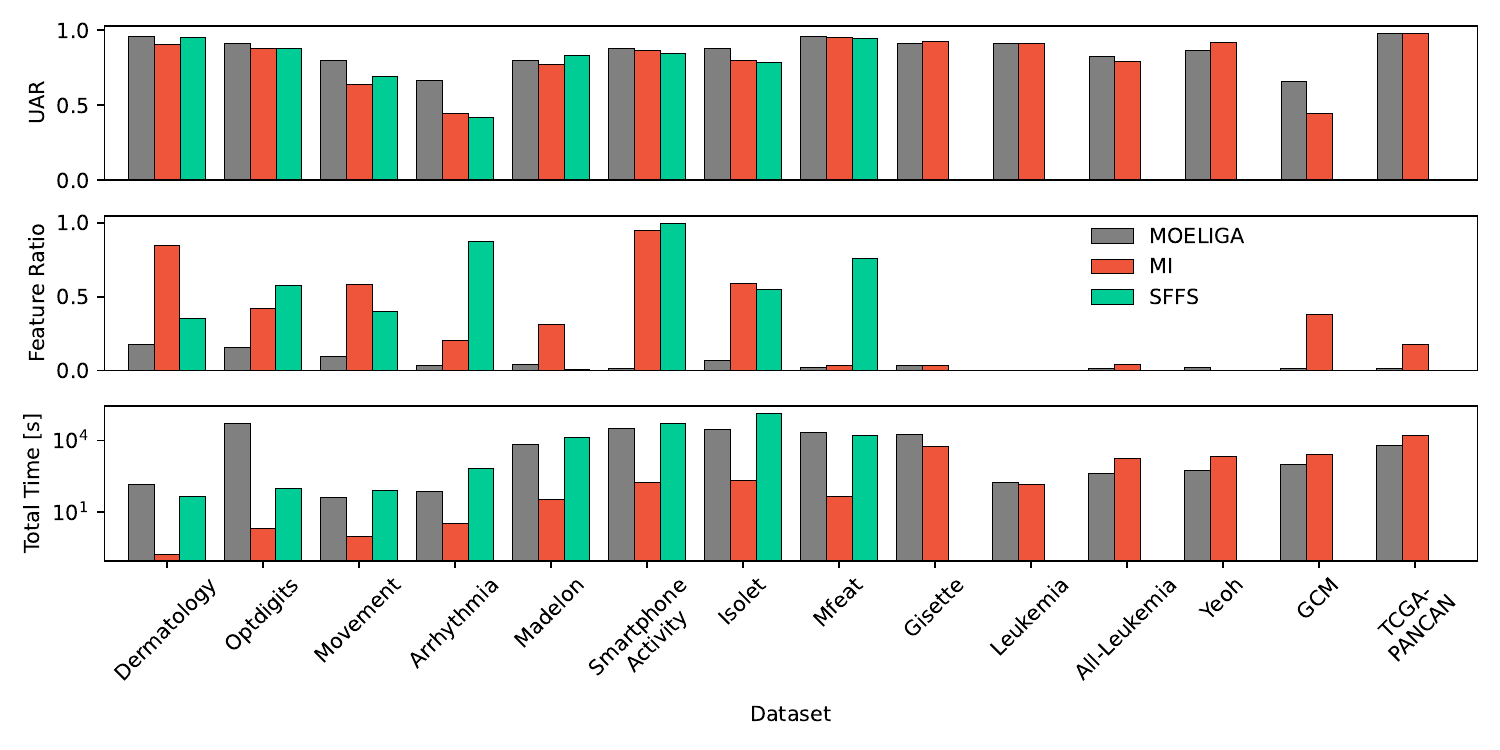} \\
    \caption{Comparison of MOELIGA, MI and SFFS in the search for optimum feature subset size.
    Top: classification performances in terms of UAR for each dataset. Gray bars correspond to MOELIGA, orange bars correspond to MI and green bars to SFFS. Center: the bars show the minimum number of features providing the maximum UAR for each method. Bottom: total elapsed time for each method. Datasets are sorted in order of increasing number of features from left to right.
    }
    \label{fig:barridosota}
\end{figure}
%%%%%%%%%%%%%%%%%%%%%%%%%%%%%%%%%%%%%%%%%%%%%%%%%%%%%%%%%%%%%%%%%%%%%%%%%%%%%%%%

An important advantage of MOELIGA is that the size of the feature subset is optimized, as well as the proper features. In contrast, all of the SOTA approaches that we have considered for comparison require the user to specify a desired number of features to select. In the previous sections, in order to compare the performace, we have set the desired feature subset size for all of the approaches to the number of features selected by MOELIGA on each dataset. In this section we present a different comparison, where the competing approaches were evaluated for all the possible number of features, in order to select their best performance by means of the UAR. Then, for each method we selected the subset size, and features, maximizing the UAR.

For this comparison, we selected two representative algorithms from the earlier evaluation: MI and SFFS, both providing competitive results across most datasets. For MI, we rank features and evaluate a DT classifier incrementally, adding one feature at a time from the top-ranked feature onward. For SFFS, we exploited its inherent property of generating subsets of all sizes from 1 to $N_s$ when targeting a subset of size $N_s$, running the algorithm once with the total number of features. This approach allowed MI and SFFS to determine their optimal subset size, enabling comparison with MOELIGA on three aspects: the computational time required for subset size optimization, the optimal subset size achieved, and the resulting UAR.

Figure \ref{fig:barridosota} compares the results obtained for MOELIGA, MI and SFFS on the $14$ datasets introduced earlier. The top bar graph compares the three approaches in terms of UAR (higher bars are better); the middle graph shows the minimum number of features providing the maximum UAR for each method (lower bars are better); and the bottom chart shows the total elapsed time for each method (lower bars are better), including the classifier training costs.

The results in Figure \ref{fig:barridosota} show that MOELIGA provides better subsets of features for most datasets in terms of classification performance with respect to UAR. Moreover, the size of the feature subsets is significantly smaller for MOELIGA in most cases. Note that in the figure, the datasets are sorted in order of increasing number of features from left to right. Then, it can be clearly seen that the improvements of MOELIGA in terms of feature subset size are more significant as the dimensionality of the data increases. Regarding computational cost, the elapsed time is longer for MOELIGA on datasets with small and moderate numbers of features, though for the cases of larger dimensionality (right side of the bar charts), MOELIGA is equally or less computationally intensive than the other methods.

These experiments show that when the size of the feature subset is also optimized, MOELIGA  provides better solutions than the other approaches in terms of both classification performance and dimensionality, with moderate computational cost.  Moreover, the computation time for MOELIGA does not rise drastically when dealing with an extremely large number of features, in contrast to other methods that become impractical in that scenario.

%-----------------------------------------

\subsection{Performance evaluation using different classifiers}

%%%%%%%%%%%%%%%%%%%%%%%%%%%%%%%%%%%%%%%%%%%%%%%%%%%%%%%%%%%%%%%%%%%%%%%%%%%%%%%%%%%%%%%%%%%%%%%%%%%%%%%%%%%%%%%%%%%%%%%%%%%

\begin{figure}[!tb]
\centering
    \includegraphics[trim = 10mm 0mm 14mm 0mm, width=0.95\textwidth]{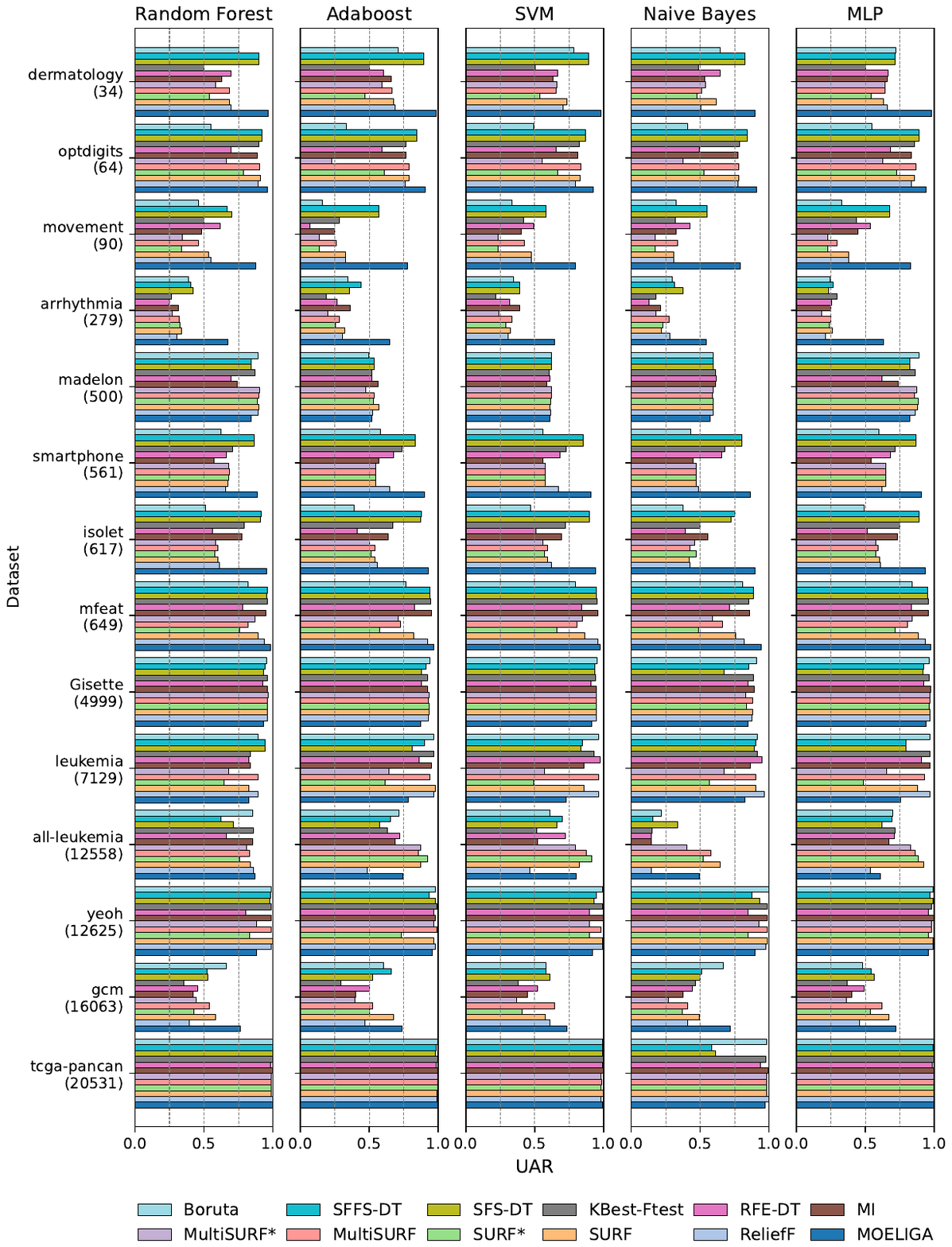}
    \caption{Comparison using different classifiers: classification performance (UAR) for MOELIGA and other FS methods.}    
    \label{fig:classifiers}
\end{figure}

%%%%%%%%%%%%%%%%%%%%%%%%%%%%%%%%%%%%%%%%%%%%%%%%%%%%%%%%%%%%%%%%%%%%%%%%%%%%%%%%%%%%%%%%%%%%%%%%%%%%%%%%%%%%%%%%%%%%%%%%%%%

Although MOELIGA can work with any classifier, it is relevant to analyze the dependency of selected features on the classifier used for fitness evaluation. To this end, we evaluated feature subsets obtained by MOELIGA using DT for fitness evaluation and by competing FS methods, testing all subsets across multiple well-known classifiers. Figure \ref{fig:classifiers} presents the results as bar charts, where each subplot corresponds to a specific classifier. Within each subplot, each group of bars correspond to a particular dataset, and the bars show the UAR achieved by the feature subset of each FS method.

The results reveal three distinct patterns. First, for 8 of 14 datasets (Dermatology, Optdigits, Movement, Arrhythmia, Smartphone-Activity, Isolet, Mfeat, and GCM), the feature subsets of MOELIGA maintain significant improvements over competing methods regardless of the classifier used. Second, for datasets Madelon, Gisette, and TCGA-PANCAN, MOELIGA achieves near-optimal performance across classifiers. Third, only for Leukemia, All-Leukemia, and Yeoh does MOELIGA exhibit appreciable performance degradation with some classifiers compared to DT.

Notably, Madelon and All-Leukemia show substantial performance variation across classifiers for all FS methods, suggesting this behavior is attributable to
dataset-specific characteristics rather than the feature selection approach. These results indicate that the multi-objective strategy of MOELIGA mitigates overfitting to specific classifiers by considering multiple objectives beyond classification performance. Consequently, feature subsets obtained with MOELIGA using DT for fitness evaluation generalize well across different classification algorithms, demonstrating the ability of the method to capture broadly relevant discriminative information.

%#############################################################
\section{Conclusions and future work}
\label{Conclusions_and_future_work}

This work presented MOELIGA, a novel multi-objective evolutionary algorithm specifically designed to address the feature selection problem while simultaneously optimizing classification performance and feature dimensionality. Through a comprehensive experimental evaluation encompassing 14 diverse datasets and comparison against 11 state-of-the-art feature selection methods, we demonstrated the efficacy and practical advantages of our proposal.

The core methodological contributions of MOELIGA include a novel crowding-based fitness sharing mechanism that promotes population diversity by accounting for cluster structures in both objective and decision spaces, a sigmoid-based transformation for the feature count objective (Eq.~\ref{eq:CRlambda}) that guides the search toward compact solutions by making cardinality differences more significant for small feature subsets, and a third objective function based on nearest neighbor distances that enhances classifier independence by evaluating feature subsets through their geometric class-separability properties. While the evolutionary local improvement strategy with subordinate populations was introduced in our previous work~\cite{eligaCLEI}, we extended and adapted it to the multi-objective context, demonstrating its effectiveness as an intensification mechanism that substantially improves solution quality without compromising diversity. The systematic sensitivity analysis conducted in two phases across 318 configurations (Section~\ref{sec:hyperparameter_effect_study}) identified optimal settings for key parameters ($\sigma = 0.0025$, $\lambda = 0.5$) and revealed that the sigmoid transformation emerged as a critical component for finding compact feature subsets.

MOELIGA achieved superior classification performance on 9 of 14 evaluated datasets, with improvements of up to 30\% in UAR compared to 
competing methods (Section~\ref{subsec:comparison_SOTA}, Table~\ref{tbl:SOTA_comparison}).
For the remaining 5 datasets, MOELIGA demonstrated competitive performance with differences below 3\% in most cases.  It is worth noting that these results were obtained while simultaneously optimizing feature subset size---a critical advantage that allows MOELIGA to automatically determine optimal dimensionality during optimization, consistently identifying substantially smaller feature subsets (often reducing dimensionality by 50--80\% in high-dimensional datasets) while maintaining or improving classification performance compared to MI and SFFS (Section~\ref{sec:barrido_sota}). The multi-objective formulation, particularly the geometry-based third objective function, promotes classifier-independent feature selection, as demonstrated by the consistently high performance of MOELIGA's feature subsets across diverse classification algorithms (Random Forest, AdaBoost, SVM, Naive Bayes, MLP), with 8 of 14 datasets showing superiority regardless of the classifier employed. Analysis of the evolved Pareto fronts (Section~\ref{subsec:pareto}) confirmed MOELIGA's ability to provide diverse trade-off solutions between classification performance and feature dimensionality, with final solutions achieving significantly higher UAR without proportionally increasing feature count.

Despite its strengths, MOELIGA has certain limitations. The computational overhead, mainly resulting from the evolutionary local improvement strategy and multiple fitness evaluations per chromosome, makes it less suitable for low-dimensional problems (fewer than 100 features), where simpler methods can achieve similar results at a much lower cost. In datasets with very limited sample sizes relative to the number of classes (e.g., Yeoh with 248 samples and 6 classes), the performance advantage of MOELIGA diminishes, possibly due to overfitting arising from repeated train-validation splits during fitness evaluation. Furthermore, although the algorithm demonstrated reasonable robustness in our experiments, it introduces several hyperparameters that may require tuning to achieve optimal performance in specific problem domains.

This work advances the state of the art in multi-objective evolutionary feature selection by introducing methodological innovations that effectively address the accuracy-dimensionality trade-off while promoting classifier independence and preventing overfitting. The comprehensive experimental evaluation demonstrates MOELIGA's practical utility, particularly for high-dimensional problems where automatic feature subset size determination provides significant value, while its ability to produce well-distributed Pareto fronts enables application-specific customization with computational efficiency competitive with existing methods for large-scale problems.

Several research directions emerge from this research. Developing adaptive mechanisms for automatically adjusting the number and frequency of subordinate population evolution based on search progress could improve efficiency while maintaining solution quality. Investigating alternative geometry-based objective functions or incorporating problem-specific knowledge could further enhance performance for challenging datasets. The integration of ensemble-based fitness evaluation, where multiple classifiers collectively guide the search, represents another promising direction for improving generalization. From a computational perspective, more sophisticated parallelization strategies beyond the current multi-core implementation could substantially reduce execution time, particularly for high-dimensional problems. Finally, extending MOELIGA to handle streaming or dynamic feature selection scenarios would broaden its applicability to real-world applications such as sensor networks or adaptive biometric systems.

%#############################################################

\section*{Acknowledgements}

The authors would like to thank the \emph{Universidad Nacional del Litoral} (with CAI+D 2024 \#85520240100155LI and CTI en Red 2025 
\#10), and the \emph{National Scientific and Technical Research Council} (CONICET), from Argentina, for their support.

%#############################################################

\section*{Declaration of generative AI and AI-assisted technologies in the man\-u\-script preparation process}

During the preparation of this work the authors used \textit{Claude} and \textit{Nature Research Assistant} in order to correct style and to improve the overall organization of the document.
After using this tools, the authors reviewed and edited the content as needed and take full responsibility for the content of the published article.

%%%%%%%%%%%%%%%%%%%%%%%%%%%%%%%%%
% \bibliographystyle{elsarticle-num}
% \bibliography{bibliography}

\begin{thebibliography}{10}
\expandafter\ifx\csname url\endcsname\relax
  \def\url#1{\texttt{#1}}\fi
\expandafter\ifx\csname urlprefix\endcsname\relax\def\urlprefix{URL }\fi
\expandafter\ifx\csname href\endcsname\relax
  \def\href#1#2{#2} \def\path#1{#1}\fi

\bibitem{cai2018}
J.~Cai, J.~Luo, S.~Wang, S.~Yang, Feature selection in machine learning: A new
  perspective, Neurocomputing 300 (2018) 70--79.
\newblock \href {http://dx.doi.org/10.1016/j.neucom.2017.11.077}
  {\path{doi:10.1016/j.neucom.2017.11.077}}.

\bibitem{mendes2020}
J.~J.~A. {Mendes Junior}, M.~L. Freitas, H.~V. Siqueira, A.~E. Lazzaretti,
  S.~F. Pichorim, S.~L. Stevan, Feature selection and dimensionality reduction:
  An extensive comparison in hand gesture classification by semg in eight
  channels armband approach, Biomedical Signal Processing and Control 59 (2020)
  101920.
\newblock \href {http://dx.doi.org/https://doi.org/10.1016/j.bspc.2020.101920}
  {\path{doi:https://doi.org/10.1016/j.bspc.2020.101920}}.

\bibitem{remeseiro2019}
B.~Remeseiro, V.~Bolon-Canedo, A review of feature selection methods in medical
  applications, Computers in Biology and Medicine 112 (2019) 103375.
\newblock \href
  {http://dx.doi.org/https://doi.org/10.1016/j.compbiomed.2019.103375}
  {\path{doi:https://doi.org/10.1016/j.compbiomed.2019.103375}}.

\bibitem{song2024}
X.~Song, Y.~Zhang, W.~Zhang, C.~He, Y.~Hu, J.~Wang, D.~Gong, Evolutionary
  computation for feature selection in classification: A comprehensive survey
  of solutions, applications and challenges, Swarm and Evolutionary Computation
  90 (2024) 101661.
\newblock \href {http://dx.doi.org/https://doi.org/10.1016/j.swevo.2024.101661}
  {\path{doi:https://doi.org/10.1016/j.swevo.2024.101661}}.

\bibitem{guyon06}
I.~Guyon, S.~Gunn, M.~Nikravesh, L.~A. Zadeh (Eds.), Feature Extraction:
  Foundations and Applications, Vol. 207 of Studies in Fuzziness and Soft
  Computing, Springer, 2006.
\newblock \href {http://dx.doi.org/10.1007/978-3-540-35488-8}
  {\path{doi:10.1007/978-3-540-35488-8}}.

\bibitem{vieira12}
S.~M. Vieira, L.~F. Mendon{\c{c}}a, G.~J. Farinha, J.~M. Sousa, Metaheuristics
  for feature selection: application to sepsis outcome prediction, in: 2012
  IEEE Congress on Evolutionary Computation, IEEE, 2012, pp. 1--8.

\bibitem{katoch2021}
S.~Katoch, S.~S. Chauhan, V.~Kumar, A review on genetic algorithm: past,
  present, and future, Multimedia Tools and Applications 80~(5) (2021)
  8091--8126.

\bibitem{labani2020}
M.~Labani, P.~Moradi, M.~Jalili, A multi-objective genetic algorithm for text
  feature selection using the relative discriminative criterion, Expert Systems
  with Applications 149 (2020) 113276.
\newblock \href {http://dx.doi.org/https://doi.org/10.1016/j.eswa.2020.113276}
  {\path{doi:https://doi.org/10.1016/j.eswa.2020.113276}}.

\bibitem{vignolo13}
L.~D. Vignolo, D.~H. Milone, J.~Scharcanski, Feature selection for face
  recognition based on multi-objective evolutionary wrappers, Expert Systems
  with Applications 40~(13) (2013) 5077--5084.
\newblock \href {http://dx.doi.org/10.1016/j.eswa.2013.03.032}
  {\path{doi:10.1016/j.eswa.2013.03.032}}.

\bibitem{hsu11}
H.-H. Hsu, C.-W. Hsieh, M.-D. Lu, Hybrid feature selection by combining filters
  and wrappers, Expert Systems with Applications 38~(7) (2011) 8144 -- 8150.
\newblock \href {http://dx.doi.org/10.1016/j.eswa.2010.12.156}
  {\path{doi:10.1016/j.eswa.2010.12.156}}.

\bibitem{pedrycz12}
W.~Pedrycz, S.~S. Ahmad, Evolutionary feature selection via structure
  retention, Expert Systems with Applications 39~(15) (2012) 11801 -- 11807.
\newblock \href {http://dx.doi.org/10.1016/j.eswa.2011.09.154}
  {\path{doi:10.1016/j.eswa.2011.09.154}}.

\bibitem{bakhshi2020}
A.~Bakhshi, S.~Chalup, A.~Harimi, S.~M. Mirhassani, Recognition of emotion from
  speech using evolutionary cepstral coefficients, Multimedia Tools and
  Applications 79~(47) (2020) 35739--35759.

\bibitem{vignolo10}
L.~D. Vignolo, H.~L. Rufiner, D.~H. Milone, J.~C. Goddard, {Evolutionary
  Splines for Cepstral Filterbank Optimization in Phoneme Classification},
  {EURASIP Journal on Advances in Signal Proc.} 2011 (2011) 8:1--8:14.

\bibitem{vignolo16}
L.~D. Vignolo, H.~L. Rufiner, D.~H. Milone, Multi-objective optimisation of
  wavelet features for phoneme recognition, IET Signal Processing 10~(6) (2016)
  685--691.

\bibitem{ghouzali2020}
S.~Ghouzali, S.~L. Marie-Sainte, Face identification based bio-inspired
  algorithms., Int. Arab J. Inf. Technol. 17~(1) (2020) 118--127.

\bibitem{khan2021}
A.~H. Khan, S.~S. Sarkar, K.~Mali, R.~Sarkar, A genetic algorithm based feature
  selection approach for microstructural image classification, Experimental
  Techniques (2021) 1--13.

\bibitem{eligaCLEI}
L.~D. Vignolo, M.~F. Gerard, Evolutionary local improvement on genetic
  algorithms for feature selection, in: 2017 XLIII Latin American Computer
  Conference (CLEI), IEEE, 2017, pp. 1--8.

\bibitem{vignolo16b}
L.~D. Vignolo, S.~M. Prasanna, S.~Dandapat, H.~L. Rufiner, D.~H. Milone,
  Feature optimisation for stress recognition in speech, Pattern Recognition
  Letters 84 (2016) 1 -- 7.
\newblock \href
  {http://dx.doi.org/http://dx.doi.org/10.1016/j.patrec.2016.07.017}
  {\path{doi:http://dx.doi.org/10.1016/j.patrec.2016.07.017}}.

\bibitem{bui1996}
T.~N. Bui, B.~R. Moon, Genetic algorithm and graph partitioning, IEEE
  Transactions on computers 45~(7) (1996) 841--855.

\bibitem{ilseok04}
I.-S. Oh, J.-S. Lee, B.-R. Moon, Hybrid genetic algorithms for feature
  selection, IEEE Transactions on Pattern Analysis and Machine Intelligence
  26~(11) (2004) 1424--1437.
\newblock \href {http://dx.doi.org/10.1109/TPAMI.2004.105}
  {\path{doi:10.1109/TPAMI.2004.105}}.

\bibitem{zhang2025}
W.~qiu Zhang, Y.~Hu, Y.~Zhang, Z.~wang Zheng, C.~Peng, X.~Song, D.~Gong, A
  multiple surrogate-assisted hybrid evolutionary feature selection algorithm,
  Swarm and Evolutionary Computation 92 (2025) 101809.
\newblock \href {http://dx.doi.org/https://doi.org/10.1016/j.swevo.2024.101809}
  {\path{doi:https://doi.org/10.1016/j.swevo.2024.101809}}.

\bibitem{rehman2024}
A.~ur~Rehman, S.~B. Belhaouari, A.~Bermak, Reinforced steering evolutionary
  markov chain for high-dimensional feature selection, Swarm and Evolutionary
  Computation 91 (2024) 101701.
\newblock \href {http://dx.doi.org/https://doi.org/10.1016/j.swevo.2024.101701}
  {\path{doi:https://doi.org/10.1016/j.swevo.2024.101701}}.

\bibitem{sivaram2019}
M.~Sivaram, K.~Batri, M.~Amin~Salih, V.~Porkodi, Exploiting the local optima in
  genetic algorithm using tabu search, Indian Journal of Science and Technology
  12~(1) (2019) 1--13.

\bibitem{navarro12}
F.~Fernández-Navarro, C.~Hervás-Martínez, R.~Ruiz, J.~C. Riquelme,
  Evolutionary generalized radial basis function neural networks for improving
  prediction accuracy in gene classification using feature selection, Applied
  Soft Computing 12~(6) (2012) 1787 -- 1800.
\newblock \href
  {http://dx.doi.org/http://dx.doi.org/10.1016/j.asoc.2012.01.008}
  {\path{doi:http://dx.doi.org/10.1016/j.asoc.2012.01.008}}.

\bibitem{thangiah2019}
S.~R. Thangiah, A hybrid genetic algorithm, simulated annealing and tabu search
  heuristic for vehicle routing problems with time windows, in: Practical
  handbook of genetic algorithms, CRC Press, 2019, pp. 347--384.

\bibitem{Sinthamrongruk2017}
T.~Sinthamrongruk, K.~Dahal, O.~Satiya, T.~Vudhironarit, P.~Yodmongkol,
  Healthcare staff routing problem using adaptive genetic algorithms with
  adaptive local search and immigrant scheme, in: 2017 International Conference
  on Digital Arts, Media and Technology (ICDAMT), 2017, pp. 120--125.
\newblock \href {http://dx.doi.org/10.1109/ICDAMT.2017.7904947}
  {\path{doi:10.1109/ICDAMT.2017.7904947}}.

\bibitem{liu2018}
X.-Y. Liu, Y.~Liang, S.~Wang, Z.-Y. Yang, H.-S. Ye, A hybrid genetic algorithm
  with wrapper-embedded approaches for feature selection, IEEE Access 6 (2018)
  22863--22874.
\newblock \href {http://dx.doi.org/10.1109/ACCESS.2018.2818682}
  {\path{doi:10.1109/ACCESS.2018.2818682}}.

\bibitem{kabir09}
M.~M. Kabir, M.~Shahjahan, K.~Murase, Involving new local search in hybrid
  genetic algorithm for feature selection, in: Int. Conf. on Neural Information
  Proc., Springer, 2009, pp. 150--158.

\bibitem{konak2006}
A.~Konak, D.~W. Coit, A.~E. Smith, Multi-objective optimization using genetic
  algorithms: A tutorial, Reliability Engineering \& System Safety 91~(9)
  (2006) 992 -- 1007.
\newblock \href {http://dx.doi.org/10.1016/j.ress.2005.11.018}
  {\path{doi:10.1016/j.ress.2005.11.018}}.

\bibitem{chen2025}
C.~Chen, X.~Yao, D.~Gong, H.~Tu, A multi-objective evolutionary algorithm for
  feature selection incorporating dominance-based initialization and
  duplication analysis, Swarm and Evolutionary Computation 95 (2025) 101914.
\newblock \href {http://dx.doi.org/https://doi.org/10.1016/j.swevo.2025.101914}
  {\path{doi:https://doi.org/10.1016/j.swevo.2025.101914}}.

\bibitem{espinosa2025}
R.~Espinosa, G.~Sánchez, J.~Palma, F.~Jiménez, Multi-objective evolutionary
  feature selection for ensemble learning with random forests in time series
  forecasting, Swarm and Evolutionary Computation 99 (2025) 102211.
\newblock \href {http://dx.doi.org/https://doi.org/10.1016/j.swevo.2025.102211}
  {\path{doi:https://doi.org/10.1016/j.swevo.2025.102211}}.

\bibitem{xue21}
Y.~Xue, H.~Zhu, J.~Liang, A.~S{\l}owik, Adaptive crossover operator based
  multi-objective binary genetic algorithm for feature selection in
  classification, Knowledge-Based Systems (2021) 107218.

\bibitem{deb14}
K.~Deb, Multi-objective optimization, in: E.~K. Burke, G.~Kendall (Eds.),
  Search Methodologies, Springer US, 2014, pp. 403--449.
\newblock \href {http://dx.doi.org/10.1007/978-1-4614-6940-7\_15}
  {\path{doi:10.1007/978-1-4614-6940-7\_15}}.

\bibitem{Rosenberg12}
A.~Rosenberg, Classifying skewed data: Importance weighting to optimize average
  recall, in: INTERSPEECH 2012, Portland, USA, 2012.

\bibitem{tang96}
K.~Tang, K.~F. Man, S.~Kwong, Q.~He, Genetic algorithms and their applications,
  {IEEE Signal Processing} 13~(6) (1996) 22--29.

\bibitem{kotsiantis2013}
S.~B. Kotsiantis, Decision trees: a recent overview, Artificial Intelligence
  Review 39~(4) (2013) 261--283.

\bibitem{curtin2013}
R.~R. Curtin, J.~R. Cline, N.~P. Slagle, W.~B. March, P.~Ram, N.~A. Mehta,
  A.~G. Gray, Mlpack: A scalable c++ machine learning library, Journal of
  Machine Learning Research 14~(Mar) (2013) 801--805.

\bibitem{bib_madelon}
I.~Guyon, {Madelon}, UCI Machine Learning Repository, {DOI}:
  https://doi.org/10.24432/C5602H (2008).

\bibitem{bib_dermatology}
N.~Ilter, H.~Guvenir, {Dermatology}, UCI Machine Learning Repository, {DOI}:
  https://doi.org/10.24432/C5FK5P (1998).

\bibitem{bib_movement}
D.~Dias, S.~Peres, H.~Bscaro, {Libras Movement}, UCI Machine Learning
  Repository, {DOI}: https://doi.org/10.24432/C5GC82 (2009).

\bibitem{bib_arrhythmia}
H.~Guvenir, B.~Acar, H.~Muderrisoglu, R.~Quinlan, {Arrhythmia}, UCI Machine
  Learning Repository, {DOI}: https://doi.org/10.24432/C5BS32 (1998).

\bibitem{bib_smartphone_activity}
J.~Reyes-Ortiz, D.~Anguita, L.~Oneto, X.~Parra, {Smartphone-Based Recognition
  of Human Activities and Postural Transitions}, UCI Machine Learning
  Repository, {DOI}: https://doi.org/10.24432/C54G7M (2015).

\bibitem{bib_isolet}
R.~Cole, M.~Fanty, {ISOLET}, UCI Machine Learning Repository, {DOI}:
  https://doi.org/10.24432/C51G69 (1994).

\bibitem{bib_optdigits}
E.~Alpaydin, C.~Kaynak, {Optical Recognition of Handwritten Digits}, UCI
  Machine Learning Repository, {DOI}: https://doi.org/10.24432/C50P49 (1998).

\bibitem{bib_gisette}
I.~Guyon, S.~Gunn, A.~Ben-Hur, G.~Dror, {Gisette}, UCI Machine Learning
  Repository, {DOI}: https://doi.org/10.24432/C5HP5B (2008).

\bibitem{bib_leukemia}
T.~R. Golub, D.~K. Slonim, P.~Tamayo, C.~Huard, M.~Gaasenbeek, J.~P. Mesirov,
  H.~Coller, M.~L. Loh, J.~R. Downing, M.~A. Caligiuri, et~al.,
  \href{https://www.kaggle.com/datasets/crawford/gene-expression}{Molecular
  classification of cancer: class discovery and class prediction by gene
  expression monitoring}, science 286~(5439) (1999) 531--537.
\newline\urlprefix\url{https://www.kaggle.com/datasets/crawford/gene-expression}

\bibitem{bib_all_leukemia}
E.-J. Yeoh, M.~E. Ross, S.~A. Shurtleff, W.~K. Williams, D.~Patel, R.~Mahfouz,
  F.~G. Behm, S.~C. Raimondi, M.~V. Relling, A.~Patel, et~al., Classification,
  subtype discovery, and prediction of outcome in pediatric acute lymphoblastic
  leukemia by gene expression profiling, Cancer cell 1~(2) (2002) 133--143.
\newblock \href
  {http://dx.doi.org/https://doi.org/10.1016/S1535-6108(02)00032-6}
  {\path{doi:https://doi.org/10.1016/S1535-6108(02)00032-6}}.

\bibitem{bib_yeoh}
K.~Sriwanna, T.~Boongoen, N.~Iam-On, Graph clustering-based discretization
  approach to microarray data, Knowledge and Information Systems 60 (2019)
  879--906.
\newblock \href {http://dx.doi.org/https://doi.org/10.1007/s10115-018-1249-z}
  {\path{doi:https://doi.org/10.1007/s10115-018-1249-z}}.

\bibitem{bib_gcm}
S.~Ramaswamy, P.~Tamayo, R.~Rifkin, S.~Mukherjee, C.-H. Yeang, M.~Angelo,
  C.~Ladd, M.~Reich, E.~Latulippe, J.~P. Mesirov, T.~Poggio, W.~Gerald,
  M.~Loda, E.~S. Lander, T.~R. Golub,
  \href{https://figshare.com/articles/dataset/S1_Data_-/21122686}{Multiclass
  cancer diagnosis using tumor gene expression signatures}, Proceedings of the
  National Academy of Sciences 98~(26) (2001) 15149--15154.
\newblock \href {http://dx.doi.org/10.1073/pnas.211566398}
  {\path{doi:10.1073/pnas.211566398}}.
\newline\urlprefix\url{https://figshare.com/articles/dataset/S1_Data_-/21122686}

\bibitem{bib_tcga_pancan}
S.~Fiorini, {gene expression cancer RNA-Seq}, UCI Machine Learning Repository,
  {DOI}: https://doi.org/10.24432/C5R88H (2016).

\bibitem{deb2007}
K.~Deb, Evolutionary multi-objective optimization without additional
  parameters, in: Parameter setting in evolutionary algorithms, Springer, 2007,
  pp. 241--257.

\bibitem{guyon2002}
I.~Guyon, J.~Weston, S.~Barnhill, V.~Vapnik, Gene selection for cancer
  classification using support vector machines, Machine learning 46~(1) (2002)
  389--422.

\bibitem{scikit}
F.~Pedregosa, G.~Varoquaux, A.~Gramfort, V.~Michel, B.~Thirion, O.~Grisel,
  M.~Blondel, P.~Prettenhofer, R.~Weiss, V.~Dubourg, J.~Vanderplas, A.~Passos,
  D.~Cournapeau, M.~Brucher, M.~Perrot, E.~Duchesnay, Scikit-learn: Machine
  learning in {P}ython, Journal of Machine Learning Research 12 (2011)
  2825--2830.

\bibitem{elssied2014}
N.~O.~F. Elssied, O.~Ibrahim, A.~H. Osman, A novel feature selection based on
  one-way anova f-test for e-mail spam classification, Research Journal of
  Applied Sciences, Engineering and Technology 7~(3) (2014) 625--638.

\bibitem{kraskov2004}
A.~Kraskov, H.~St{\"o}gbauer, P.~Grassberger, Estimating mutual information,
  Physical review E 69~(6) (2004) 066138.

\bibitem{ross2014}
B.~C. Ross, Mutual information between discrete and continuous data sets, PloS
  one 9~(2) (2014) e87357.

\bibitem{pudil1994}
P.~Pudil, J.~Novovičová, J.~Kittler, Floating search methods in feature
  selection, Pattern Recognition Letters 15~(11) (1994) 1119--1125.
\newblock \href
  {http://dx.doi.org/https://doi.org/10.1016/0167-8655(94)90127-9}
  {\path{doi:https://doi.org/10.1016/0167-8655(94)90127-9}}.

\bibitem{ververidis2008}
D.~Ververidis, C.~Kotropoulos, Fast and accurate sequential floating forward
  feature selection with the bayes classifier applied to speech emotion
  recognition, Signal Processing 88~(12) (2008) 2956--2970.
\newblock \href
  {http://dx.doi.org/https://doi.org/10.1016/j.sigpro.2008.07.001}
  {\path{doi:https://doi.org/10.1016/j.sigpro.2008.07.001}}.

\bibitem{raschkas2018}
S.~Raschka, Mlxtend: Providing machine learning and data science utilities and
  extensions to python’s scientific computing stack, The Journal of Open
  Source Software 3~(24).
\newblock \href {http://dx.doi.org/10.21105/joss.00638}
  {\path{doi:10.21105/joss.00638}}.

\bibitem{kursa2010a}
M.~B. Kursa, A.~Jankowski, W.~R. Rudnicki, Boruta--a system for feature
  selection, Fundamenta Informaticae 101~(4) (2010) 271--285.

\bibitem{breiman2001}
L.~Breiman, Random forests, Machine learning 45~(1) (2001) 5--32.

\bibitem{kursa2010b}
M.~B. Kursa, W.~R. Rudnicki, et~al., Feature selection with the boruta package,
  J Stat Softw 36~(11) (2010) 1--13.

\bibitem{kononenko1996}
I.~Kononenko, M.~Robnik-Sikonja, U.~Pompe, Relieff for estimation and
  discretization of attributes in classification, regression, and ilp problems,
  Artificial intelligence: methodology, systems, applications (1996) 31--40.

\bibitem{Urbanowicz2017}
R.~J. Urbanowicz, R.~S. Olson, P.~Schmitt, M.~Meeker, J.~H. Moore, Benchmarking
  relief-based feature selection methods, arXiv e-print.
  https://arxiv.org/abs/1711.08477 (2017).

\bibitem{greene2009}
C.~S. Greene, N.~M. Penrod, J.~Kiralis, J.~H. Moore, Spatially uniform relieff
  {(SURF)} for computationally-efficient filtering of gene-gene interactions,
  BioData mining 2~(1) (2009) 1--9.

\bibitem{greene2010}
C.~S. Greene, D.~S. Himmelstein, J.~Kiralis, J.~H. Moore, The informative
  extremes: using both nearest and farthest individuals can improve relief
  algorithms in the domain of human genetics, in: European conference on
  evolutionary computation, machine learning and data mining in bioinformatics,
  Springer, 2010, pp. 182--193.

\bibitem{granizo2013}
D.~Granizo-Mackenzie, J.~H. Moore, Multiple threshold spatially uniform
  {ReliefF} for the genetic analysis of complex human diseases, in: European
  conference on evolutionary computation, machine learning and data mining in
  bioinformatics, Springer, 2013, pp. 1--10.

\bibitem{Derrac2011}
J.~Derrac, S.~García, D.~Molina, F.~Herrera, A practical tutorial on the use
  of nonparametric statistical tests as a methodology for comparing
  evolutionary and swarm intelligence algorithms, Swarm and Evolutionary
  Computation 1~(1) (2011) 3--18.
\newblock \href {http://dx.doi.org/10.1016/j.swevo.2011.02.002}
  {\path{doi:10.1016/j.swevo.2011.02.002}}.
  
\bibitem{Demsar2006}
Dem\v{s}ar, J.
\newblock Statistical Comparisons of Classifiers over Multiple Data Sets.
\newblock {\em Journal of Machine Learning Research}, 7(1):1--30, 2006.
\newblock URL: \url{http://jmlr.org/papers/v7/demsar06a.html}

\end{thebibliography}

\end{document}